\documentclass[review,3p,times,11pt]{elsarticle}

\usepackage{enumitem}
\usepackage{lineno}
\usepackage[colorlinks,linkcolor=red,anchorcolor=blue,citecolor=green]{hyperref}
\usepackage{multirow}
\usepackage{rotating}
\usepackage{epstopdf}
\usepackage{graphicx}
\usepackage{subfigure}

\usepackage{epstopdf}
\usepackage{times}
\usepackage{longtable}

\usepackage{makeidx}
\usepackage{multirow}
\usepackage{multicol}
\usepackage[dvipsnames,svgnames,table]{xcolor}
\usepackage{graphicx}
\usepackage{epstopdf}
\usepackage{ulem}
\usepackage{url}

\usepackage{array}
\usepackage{amsmath,amssymb,amsfonts}
\usepackage{algorithmic}
\usepackage{algorithm}
\usepackage{graphicx}
\usepackage{textcomp}

\usepackage{epsfig}
\usepackage{makecell}
\usepackage{longtable,booktabs}
\usepackage{longtable}

\usepackage{lscape}

\usepackage[export]{adjustbox}

\usepackage{natbib}
\usepackage[skip=0.333\baselineskip]{caption}

\usepackage{placeins}
\usepackage{caption}

\journal{Journal}

\begin{document}
\begin{frontmatter}
	
	
		
	\title{Knowledge-infused Deep Learning Enables Interpretable Landslide Forecasting}


\author[label1]{Zhengjing Ma}
\author[label1]{Gang Mei\corref{cor1}}
\ead{gang.mei@cugb.edu.cn}
\cortext[cor1]{Corresponding author}
\address[label1]{School of Engineering and Technology, China University of Geosciences (Beijing), 100083, Beijing, China}

\begin{abstract}
	
Forecasting how landslides will evolve over time or whether they will fail is a challenging task due to a variety of factors, both internal and external. Despite their considerable potential to address these challenges, deep learning techniques lack interpretability, undermining the credibility of the forecasts they produce. The recent development of transformer-based deep learning offers untapped possibilities for forecasting landslides with unprecedented interpretability and nonlinear feature learning capabilities. Here, we present a deep learning pipeline that is capable of predicting landslide behavior holistically, which employs a transformer-based network called LFIT to learn complex nonlinear relationships from prior knowledge and multiple source data, identifying the most relevant variables, and demonstrating a comprehensive understanding of landslide evolution and temporal patterns. By integrating prior knowledge, we provide improvement in holistic landslide forecasting, enabling us to capture diverse responses to various influencing factors in different local landslide areas. Various nonlinear activation mechanisms in the architecture facilitate mapping prior knowledge and data into a high-dimensional representation space, enabling high-level pattern recognition. Therefore, it can effectively capture the intricate interdependencies between internal and external factors contributing to landslides. Moreover, interpretability here can be instrumental in identifying temporal patterns in landslide behavior and determining key predictors. Combining an attention mechanism and adaptive feature selection enables our approach to reveal the underlying dynamics of landslides, leading to a comprehensive understanding of trends and influencing factors that influence landslide behavior. Using deformation observations as proxies for measuring the kinetics of landslides, we validate our approach by training models to forecast reservoir landslides in the Three Gorges Reservoir and creeping landslides on the Tibetan Plateau. When prior knowledge is incorporated, we show that interpretable landslide forecasting effectively identifies influential factors across various landslides. It further elucidates how local areas respond to these factors, making landslide behavior and trends more interpretable and predictable. The findings from this study will contribute to understanding landslide behavior in a new way and make the proposed approach applicable to other complex disasters influenced by internal and external factors in the future.

\end{abstract}

\end{frontmatter}



%
%
%
%
%
%
%
%


\section{Introduction}
\label{sec1}

As one of the most common natural hazards worldwide, landslides have caused considerable economic losses and fatalities \cite{001Froude20182161,002Petley2012927}. These events are often followed by cascading effects, including damming and flooding, which exacerbate the vulnerability of the affected regions \cite{003Fan2019421,004Yang2022}. Further highlighting the importance of addressing landslide hazards is the trend of accelerating urbanization coupled with the threat of climate change \cite{005Patton2019116,006Dille20221048}. It is essential to design and implement effective risk reduction strategies to mitigate the effects of landslides on economies, public safety, and local ecosystems \cite{007Gariano2016227,008Piciullo2018228}. This requires an enhanced ability to predict (forecast, project, and anticipate) landslides \cite{009Guzzetti}. Research has demonstrated that different types of landslides exhibit distinct kinematic behaviors, which are influenced by various factors, such as precipitation, seismic events, glacial retreat, and human activity \cite{010Hungr2014167,011Guzzetti20083,012Kargel2016,013Lacroix2022}. Due to the difficulty of directly observing landslides, deformation measurements, including displacement time series, are typically regarded as a proxy to determine their underlying kinematics \cite{014Hu2020,015Frattini20181053}. The measurements of landslide deformation, obtained through a wide variety of slope monitoring techniques, are critical to advancing the scientific understanding of landslide deformation and its interaction with influencing factors \cite{016Zeng2023,017Hu2021}. Analysis of slope deformation-related measurements provides an opportunity to establish patterns and trends in landslide behavior, helping to uncover the mechanisms triggering landslides and enabling advanced forecasting models to be developed \cite{018Lei2023,019Cascini20222839}.

Forecasting landslides involve two paradigms: geotechnical models and phenomenological approaches \cite{020Bru2018257,021Zhou2022}. Models based on geotechnical principles shed light on the underlying mechanisms that govern landslide instability and temporal progression. Even though these physical-based models have significant advantages, they are often limited to specific cases, which hinders their widespread application. The complex, non-linear relationships between dynamic variables in natural environments can also be challenging to understand and measure, limiting the ability of these models to capture all relevant physical processes. In phenomenological approaches, especially data-driven models, correlations are established between input variables, such as rainfall, and output variables, such as landslide displacements \cite{022Huang2017173,023Intrieri20162501}. Complex interactions can be efficiently modeled without explicitly addressing the underlying physical mechanism by using these data-driven models, which are dependent on observational data \cite{024Guo2020}. The problem is that they do not consider a system's internal workings and therefore are referred to as black-box models. There are distinct advantages and limitations associated with both methods. Geotechnical models provide insight, but they are difficult to generalize. Phenomenological models provide a practical approach to prediction at the expense of scientific understanding \cite{025Zou2020}.

Recent advancements in high-resolution monitoring technologies have resulted in an unprecedented increase in landslide behavior measurement, particularly deformation \cite{026Intrieri20171713,027Casagli202351}. The availability of large amounts of data has facilitated the development of data-driven approaches to analyze and forecast landslides \cite{028Wasowski2020445,029Bekaert2020}. This has led to a growing interest in analyzing and forecasting landslide deformation, as it provides valuable insights into landslides' behavior under the influence of multiple factors. Artificial intelligence, specifically machine learning models, has played a crucial role in accurately forecasting landslide deformations using displacement measurements \cite{030Nava2023}. These models have proven highly effective, particularly in cases where deformations exhibit periodic variability, enabling accurate mid-to-long-term forecasts. Various machine learning approaches, including fuzzy logic algorithms and support vector machines, have demonstrated success in this filed \cite{031Wang2022,032Miao2018475,033Wen20172181,034Deng2021,035Long2022,036Ma20222489}. Furthermore, deep learning methods, such as convolutional and recurrent neural networks, have improved performance in forecasting landslide deformations \cite{037Yang2019677,038Pei20217403}[37, 38]. The superior performance of deep learning methods can be attributed to their intrinsic ability to identify complex nonlinearities, model intricate relationships between different input features, and learn hidden patterns present in the input \cite{039Lusch2018,040Mousavi2022}.

However, machine learning-based approaches for landslide deformation forecasting encounter a significant limitation as they heavily rely on displacement measurements from individual monitoring points, resulting in a narrow focus that impedes a comprehensive understanding of the complex interplay among geological, meteorological, and terrain factors governing landslide behavior. Recent studies have aimed to overcome the limitations of data-driven models in landslide deformation forecasting by exploring the integration of additional influential factors \cite{041Liu2020}. This focus aims to enhance forecast accuracy and deepen our understanding of landslide deformation evolution at specific monitoring points. While considerable attention has been given to incorporating time-dependent environmental observation variables, the effective integration of static information and time-varying information derived from field investigations or historical records remains an underexplored area, presenting challenges for holistic landslide forecasting. To address this research gap, there has been a growing recognition of the value of integrating domain knowledge and expert insights into the modeling process \cite{041Liu2020,042Guo2022,043zhang2023skilful,044Han2021}. While previous studies have considered various internal characteristics, such as lithological composition and specific monitoring areas, in landslide susceptibility mapping to predict landslide occurrences, the forecasting of landslide dynamics has not adequately leveraged this prior knowledge. Integrating static knowledge, alongside time-varying data from environmental observations or field investigations, into neural networks holds excellent potential for improving their understanding of landslide behavior's temporal and spatial aspects.

In addition to the aforementioned advancements, recent efforts have concentrated on integrating interpretable machine-learning techniques with domain knowledge to bridge the gap between complex models and human understanding \cite{045Youssef2023,046Mondini2023,047Li202229}. The lack of algorithmic transparency in contemporary machine learning methods challenges their interpretability, as their underlying mechanisms remain elusive. Given the high-stakes nature of landslide prediction, where decisions directly impact human lives and involve significant costs associated with insurance and reconstruction, numerous studies have emphasized the importance of exercising caution when selecting appropriate artificial intelligence models for specific landslide deformation forecasting scenarios \cite{048Lian201591}. Recent research advocates for developing transparent and interpretable machine learning models for landslide forecasting, aiming to decipher the decision-making processes employed by complex models \cite{049Dahal2023,050Sun2022}. Unraveling these processes can yield insights into the factors and patterns influencing landslide deformation.

Consequently, incorporating domain knowledge into interpretable machine learning or deep learning has substantial implications for fostering trust and understanding in landslide forecasting results. Various approaches have emerged in the pursuit of interpretability, such as rule-based models, decision trees, and attention mechanisms in deep learning architectures. These techniques facilitate the extraction of meaningful features and provide explanations for model predictions. Attention mechanisms, or similar approaches, offer promising prospects for integrating domain knowledge into interpretable deep neural networks, ultimately contributing to establishing trust and understanding in landslide forecasts. One notable attention-based neural network is transformers. Indeed, transformers have gained significant attention for their exceptional performance in natural language processing tasks. However, their applicability extends beyond language-related tasks \cite{051Vaswani}. Transformers have shown promise in various sequence modeling tasks, including time series forecasting \cite{052Wen2022TransformersIT}. The inherent attention mechanisms in transformers enable interpretability by explicitly highlighting influential factors and contributing features throughout the forecasting process. The application of transformers in landslide forecasting offers several unique advantages. First, transformers can capture local and global dependencies within the data, allowing for a comprehensive understanding of landslide dynamics. Their self-attention mechanism empowers them to assign different weights to various elements in the input, effectively emphasizing the most influential factors driving landslide deformation. Moreover, the attention mechanisms in transformers can be extended to incorporate domain knowledge and expert insights. With prior knowledge, the model can focus on specific factors, regions of interest, and landslide behavior at a particular time step, thereby enhancing the predictability and interpretability of landslide evolution.

Herein, we achieve knowledge-guided Interpretable Landslide Forecasting by infusing prior knowledge into a transformer-based neural network named Landslide Forecasting Interpretable Transformer (LFIT). LFIT incorporates a variable selection mechanism and an attention mechanism within the transformer architecture, which enables the weights to be extracted and visualized, which provides valuable insights into network dynamics. These weights represent the non-linear characteristics that the neural network learns, which enables a fundamental understanding of which input features and landslide behavior at different time points are focused on in the model, highlighting their relative importance. Our extensive experiments employed two types of landslides to evaluate LFIT's effectiveness. This study shows that LFIT identifies and reveals significant temporal patterns and features critical for accurate landslide forecasting by integrating landslide property information specific to certain monitoring points. By capturing the temporal patterns in local areas that respond to various influencing factors during landslide evolution, LFIT enhances its ability to extrapolate future landslide deformation trends and improves overall predictability in landslide forecasting. Specifically, our approach incorporates static information on landslide properties as prior knowledge into LFIT, enhancing the model's spatiotemporal awareness and capacity to capture the underlying dynamics of landslide processes. By integrating static information, we can understand the variables that contribute to the behavior of landslides. Considering geological characteristics, LFIT has gained a deeper understanding of the complex dynamics involved in transitioning from a creeping landslide to a catastrophic event. A further benefit of LFIT is its ability to capture the intricate interaction between influencing factors and deformations both at a local and a holistic level, allowing for the identification of critical factors essential for accurate and comprehensive landslide forecasting in particular places. Particularly, LFIT demonstrated exceptional performance in the case of reservoir landslides, where it can effectively learn from deformation data near the water source. LFIT successfully detects and elucidates cyclical patterns associated with reservoir fluctuations. LFIT identifies water level as the most influential factor in these cases, leading to highly accurate long-term predictions. The significant findings of this study highlight the effectiveness of our knowledge-guided approach in landslide deformation forecasting. LFIT produces physically realistic and consistent results by incorporating prior knowledge. The interpretability of LFIT's results and its ability to capture temporal dynamics and interactions among influencing factors opens new avenues for understanding landslide behavior and improving landslide predictability.

\section{Methods}
\label{sec:2:methods}
\subsection{Data and knowledge acquisition}
\label{subsec:2:1:Data and knowledge acquisition}
Our interpretable landslide forecasting pipeline has been evaluated on two distinct types: reservoir-induced and large-scale landslides. These types exhibit unique deformation characteristics. Reservoir landslides are typically characterized by a distinctive step-like variation in displacement that are measured by in-situ devices \cite{053Yao201934}. This step-like pattern indicates sudden and discrete movements triggered by specific events or conditions.

Fluctuations in the reservoir water level, intense rainfall, or changes in the seepage field play pivotal roles in influencing the displacement behavior of reservoir landslides \cite{054Zou2023}. Understanding the underlying mechanisms behind these step-like variations is essential for understanding the dynamic nature of reservoir landslides and developing more accurate forecasting models \cite{055Yang2023}. Furthermore, creeping landslides, especially those that move slowly, typically exhibit a gradual progression that ultimately leads to a sudden and catastrophic collapse \cite{056Lacroix2020404,057Liu2020}. Investigating deformation patterns derived from measurements of their creeping displacement can reveal signs of potential acceleration or failure, essential for developing effective early warning systems and proactive management strategies \cite{058Murphy2022,059UrgilezVinueza20222233}.

Our case investigation initially concentrates on reservoir-induced landslides in the Three Gorges reservoir area, which is highly susceptible to landslides owing to a confluence of factors such as a rainy climate, geology predisposed to sliding, and widespread human activity. Prior investigations have identified this region's primary triggers of landslides: precipitation and reservoir water level fluctuations \cite{060Ye2022}. The periodic variations in the reservoir water level have significant consequences, as they alter the seepage field and the stress state of the landslide mass. Elucidating the complex interplay between water level changes and landslide behavior is crucial for comprehending and forecasting landslide behavior in the area.

We collected diverse time series data from the monitoring site in one of the landslides named Huanglianshu landslide, located in Fengjie County, Chongqing, on the right bank of the Yangtze River. It has a circular-chair-shaped distribution in plan view, with a slope direction of $350^{\circ}$ and slope angles between $5^{\circ}$ and $30^{\circ}$. The landslide spans approximately 700 meters in length and 650 meters in width, with a height difference of about 120 meters between the front and rear edges. The sliding material is clay with fragmented rocks, thicker in the center and thinner on the sides. The loose surface structure allows for easy infiltration of precipitation. The landslide can be divided into three parts based on composition and permeability:(1)The upper part with gravelly soil of high permeability.(2) The middle part with cohesive and interbedded gravelly soil of poor permeability.(3) The lower part with thin layers of sand and clay.

The lower section has experienced erosion and softening due to reservoir water, leading to crack formation. During heavy rainfall, these cracks expand, increasing the landslide's permeability. The combined effects of rainfall infiltration and a decrease in reservoir water level contribute to an emergent event. Huanglianshu landslide's leading edge was severely deformed on May 31, 2012.

We obtained long-term monitoring data (cumulative displacements) for the Huanglianshu landslide from in situ measurements conducted between March 1, 2005, and June 1, 2012. The dataset includes nine monitoring points with the monthly resolution, hourly rainfall, and monthly reservoir level information. We classified the monitoring points into danger, non-danger, and near-danger areas based on qualitative zonings provided in the landslide documents. We also considered the dominant soil types within the study area, specifically cohesive and gravelly soil. The monitoring site itself represents a local patch of landslide deformation, with designated names for each monitoring point: FJ-004, FJ-005, FJ-006, FJ-008, FJ-009, FJ-010, FJ-012, FJ-013, and FJ-014. The original records of the Huanglianshu landslide monitoring data contain many missing values. To mitigate the potential impact of these missing values on the effectiveness of the proposed method, monitoring point information with missing values exceeding 70\% was removed.

Our second case study focuses on large landslides on the Qinghai-Tibet Plateau, an area with extensive tectonic activity, mega-earthquakes, glaciers, high stress, and deeply incised valleys. These adverse geological and climatic conditions in this region lead to a highly complex mechanism of landslides, in which internal geological factors and external triggers include seasonal precipitation, regional geology, earthquakes, and high stress \cite{061Zhang20213577}. Upon exposure to intense weathering, the slope surfaces are compromised regarding mechanical properties and integrity. Thus, these slopes are susceptible to external driving forces, further exacerbating adverse geological conditions. We collected data from the Baige Landslide, which is representative of the region's large landslides. Located in the upper Jinsha River region of the southern Tibetan Plateau, this landslide has been extensively studied in recent decades \cite{057Liu2020,064Fan20191003,064Fan20191003}. The elevation in this area ranges from 2890 to 3720 meters, with a relative height difference of 830 meters. In the aftermath of two consecutive landslides in 2018, several cracks were observed behind the prominent head scarp, revealing three steep free surfaces caused by stress redistribution. There are sliding blocks present in the landslide source area. An extensive investigation has identified three major deformation zones: K1, K2, and K3 \cite{064Fan20191003}. An array of global navigation satellite systems (GNSS) have been deployed in order to monitor the current movement of the Baige landslide, as well as sensors for measuring the temperature and humidity of the exterior environment. Besides the K1, K2, and K3 zones, the monitoring also covers the slope on the southern side of the K2 zone, referred to as the K4 zone.

Landslide monitoring records from the Baige Landslide are publicly available with different temporal resolutions, such as 3 hours and 24 hours. The data quality has been improved by removing any displacement records with significant missing data from specific monitoring points. Since the employed model only considers displacement time series of similar lengths and temporal resolutions, landslide displacement time series corresponding to the same period and resolution are retained. It's worth noting that the three temporal resolution landslide records cover different monitoring points and periods, as detailed in the supplementary information. Data collected from each monitoring point consists of (1) continuous observations of cumulative displacement in horizontal and vertical directions and (2) the point's coordinates in real time. Data collection for this study began in 2019. These landslide displacement records provide an opportunity to investigate how landslides evolved after two significant hazard events in 2018 and forecast future deformation patterns and future devastation caused by the Baige landslide. The August 1, 2022, date is the endpoint for all Baige landslide monitoring datasets.

\subsection{Predictor and predictand variable }
\label{subsec:2:2:Predictor and predictand variable}

In the reservoir landslide case, the significant predictors consist of the following:(1) Time-varying displacement measurements were obtained from multiple monitoring points.(2)Measurements of the water level and precipitation were collected through ground instruments.(3)Categorical information encompassing the degree of danger, soil type, and monitoring site labels.

Predictand variables are displacement measurements. Please note that displacement time series is a combination of multiple component measurements. The recorded cumulative displacements $\boldsymbol{x}$ in $\boldsymbol{k}$ directions at each monitoring point were integrated by $\|x\|_2:=\left(\sum_{k=1}^K\left|x_k\right|^2\right)^{1 / 2}$, thus reflecting the overall spatial state of landslide displacement. 

For the representative landslide in the Qinghai-Tibet Plateau, the primary predictors include the following:(1) Time-varying displacement measurements from multiple monitoring points.(2) Ground instrument measurements of precipitation, temperature, and humidity.(3)Information on the unstable region and monitoring site labels.

The predictand variables are displacement measurements in both horizontal and vertical directions. For the Baige landslide, we do not calculate the sum of squares of horizontal and vertical displacements, but we retain them as a result. This decision was made because the horizontal and vertical displacements reflect deformation characteristics at different monitoring locations. These are expected to elucidate the interaction of various positions on the landslide body.

These variables were selected after an extensive review of previous literature and an intensive data mining effort. According to the previous study, several significant factors contributed to the Huanglianshu landslide. For instance, the fluctuations in reservoir water levels lead to periodic unloading dynamics of the landslide displacement, with distinct step changes occurring in response to reservoir water unloading. The Huanglianshu landslide was triggered by heavy rainfall in 2012. There was a tendency for water infiltration to occur more readily in the danger zone of the landslide located between the middle and upper parts, which was characterized by predominant gravel grain sizes and loose surface structures. After heavy rainfall in 2007, substantial variation was observed around the monitoring points. Additionally, heavy rain in May 2012 resulted in the development of shear tension cracks on both sides of the slope body edge, transverse tension cracks in the middle and upper parts, and drum tension cracks on the front edge, ultimately triggering a critical slip alert for the landslide.

Potentially unstable zones in baige landslides are more susceptible to instability under adverse conditions, particularly in the event of heavy rain. Tectonics and Jinsha River downcutting produced steep terrain with high peaks and deep V-valleys, adverse geology for landslides. High water enhances flow, adversely affects geology, and enables easier infiltration and accelerated erosion. Temperature and humidity data provide climate change/extreme weather context; previously stable slopes may slide as the ice melts rapidly. The "K1, K2, K3, K4" zone classification designates distinct categories for specific landslide sections, providing detailed descriptions of the surrounding conditions and encapsulating prior knowledge. Such classifications have geological significance, and each embodies local conditions. e.g., K1 has gneiss/schist, arc cracks at multiple scales; K2 has highly weathered carbonaceous slate, arc cracks at multiple scales; K3 has phyllite without serpentinite, with gneiss below, and has distributed cracks \cite{064Fan20191003}.

\subsection{ Incorporating Prior Knowledge into an Interpretable Deep Neural Network Architecture: A Methodological Framework}
\label{subsec:2:3:Incorporating Prior Knowledge into an Interpretable Deep Neural Network Architecture}

\begin{figure*}[!ht]
	\centering
	\captionsetup{labelfont=bf}
	\includegraphics[width=16.5cm]{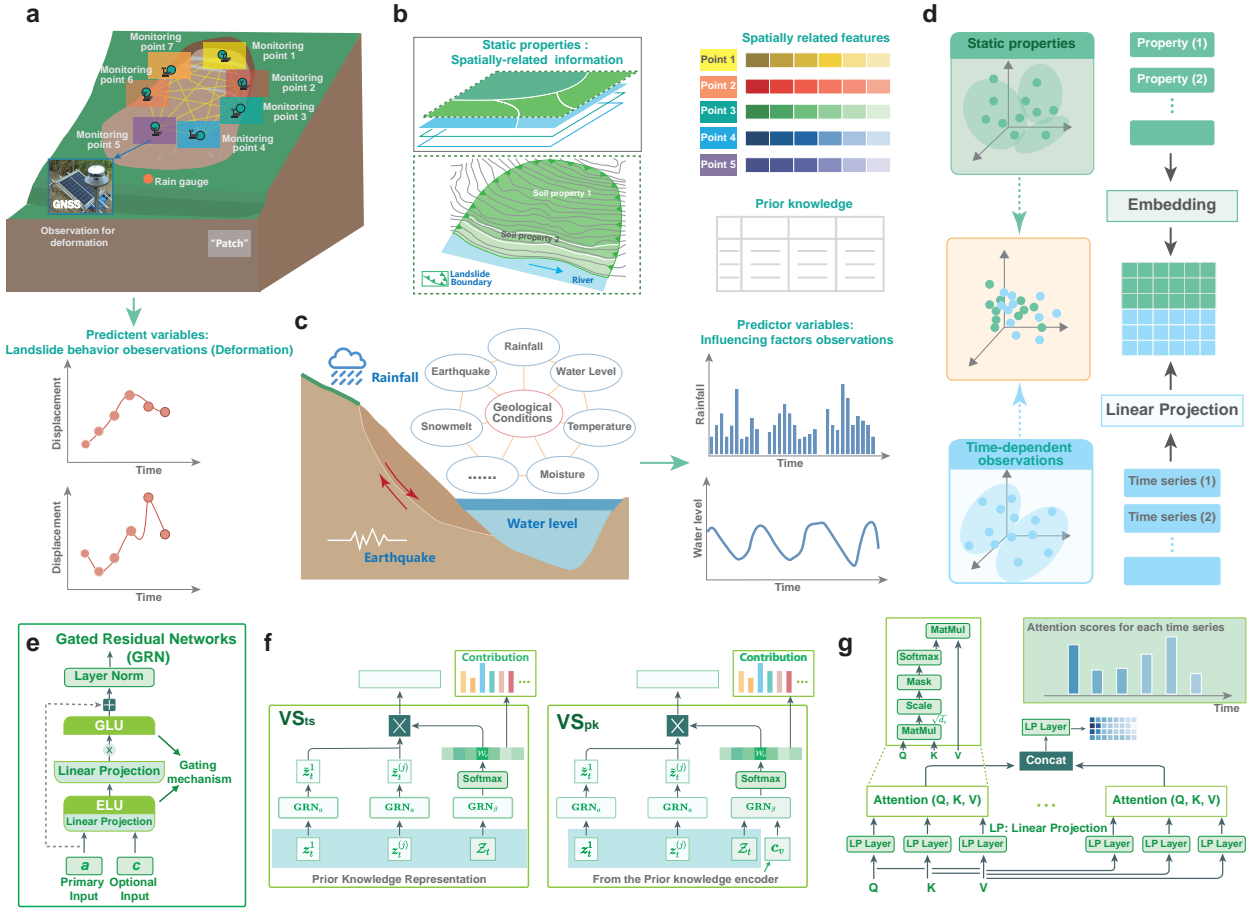}
	\caption{\textbf{Schematic view of infusing Knowledge into Interpretable Deep Learning for Landslide Forecasting}. \textbf{a}. The observations of scattered observation points are governed by the integral landslide system with complex spatiotemporal correlations. We introduce the concept of "Patches" to represent localized conditions in the vicinity of each observation point. The interconnections between these Patches capture the overall system behavior of the landslide. \textbf{b}. Exploiting spatially relevant information as a priori knowledge to characterize landslide properties in proximity to observation points. \textbf{c}. The intricate interplay between internal and external factors affecting landslide dynamics and the observation of these dynamic influences. \textbf{d}. Embedding a priori knowledge and observed time series into a high dimensional Space for Mapping. \textbf{e-g}. Three key components of interpretable deep neural network LFIT: GRU(\textbf{e}), VS(\textbf{f}), interpretable attention block (\textbf{g}).} 
	\label{Figure1}
\end{figure*}

LFIT, a deep neural network tailored for interpretable landslide forecasting, builds upon the Temporal Fusion Transformer architecture, a deep learning variant celebrated for its exceptional forecasting performance \cite{,065Lim20211748}. The foundation of these functional blocks is inspired by the transformer model, originally proposed by the Google research team in 2017. Its distinct architecture allows for the seamless flow of meaningful information by utilizing mappings and nonlinear transformations. Fig. \ref{Figure1} provides a detailed description of the LFIT architecture, which comprises several fundamental components commonly found in deep neural networks, including feed-forward layers, LSTM units, residual connections, and gating mechanisms. These elements are organized into functional blocks designed to perform specific tasks, such as variable selection, LSTM-based encoder-decoder structures, and interpretable attention mechanisms. Further details regarding the neural network components can be found in Table 1.

onsidering the complexity of landslide data and knowledge, non-linear modeling techniques are required to completely capture the drivers of landslide behavior and the impacts suffered. Multiple gated residual networks (GRN) blocks are incorporated into LFIT to identify and leverage the most relevant features and relationships across multi-modal inputs. This block contains two types of gating mechanism: Exponential Linear Units (ELU) and Gated Linear Units (GLU). ELU activation involves removing irrelevant information through a nonlinear process. Specifically, $\mathrm{ELU}(\mathrm{x})=\{x, x>0 ; \alpha(\exp (x)-1), x \leq 0\}$, where parameter $\alpha$ determines the negative saturation value. Due to their smoothness and nonlinearity, ELUs can push mean unit activations closer to zero, thereby speeding up learning and resolving sharp smoothing problems incurred by rectified linear units (ReLUs) and other commonly used activations. 

GLU performs a component-wise product of linear transformations, one of which is Sigmoid-activated. A GLU performs: $\operatorname{GLU}(x)=\left(x * \Theta_1+\mathbf{b}_1\right) \odot \sigma\left(x * \Theta_2+\mathbf{b}_2\right)$. Where $\Theta_1$ and $\mathbf{b}_1$ are the weight matrix and bias of the first linear transformation, $\Theta_2$ and $\mathbf{b}_2$ are the weights and bias of the second transformation which is activated by a sigmoid function $\sigma(\cdot)$, and $\odot$ is the element-wise product. The Sigmoid activation limits the output to 0 and 1, allowing us to focus on the most relevant information while suppressing irrelevant information, thereby helping to slow down the impact of unhelpful features. By implementing dropout and layer normalization techniques, generalization and convergence are facilitated, thereby preventing overreliance on original data. A residual connection is also incorporated into the GRN, which directly extracts the remaining information features from preceding layers, thus combining linear and nonlinear contributions from GLU. Understanding linear and nonlinear relationships is crucial to understanding landslide dynamics.

Leveraging the robust nonlinear projection capabilities of GRNs, our proposed methodology seamlessly incorporates prior knowledge into the LFIT framework through a three-step process.

\subsubsection{ Step 1: Integrating Prior Knowledge and Data using High-Dimensional Embeddings}
\label{subsection 3:3:1 Step 1: Integrating Prior Knowledge and Data using High-Dimensional Embeddings}

In the initial layer of LFIT, as outlined in Table 1, a diverse array of input variables undergoes a transformation process to generate input embeddings. Each information channel is linked to a distinct learnable projection, enabling the mapping of data and prior knowledge into high-dimensional spaces that can be effectively processed and integrated by deep learning models. 

A linear projection maps the time-dependent landslide deformation or driver variable into a $d_m$ dimensional representation. Simultaneously, we incorporate qualitative knowledge from domain experts describing regional landslide hazards and geotechnical properties. This knowledge is encoded into high-dimensional spaces using word embeddings to capture latent relationships between geology and landslides. Specifically, we obtain relevant prior knowledge from documents and expert insights describing the geological conditions surrounding each monitoring site. We map this categorical information into a $d_m$ dimensional representation, aligning it with the same high-dimensional space used for other input variables, where $d_m$ denotes the dimension that can be customized based on specific requirements. The embedding representations serve as high-level features that offer insights into the complex behaviors of landslides under static properties and dynamic observations.

By unifying all relevant information into this shared space, LFIT ensures that time-dependent and descriptive static information is effectively captured and integrated. Following the transformation process, each $j$-th input variable is mapped into a vector $\mathbf{z}_t^{(j)} \in \mathbb{R}^{d_m}$. This step serves as the foundation for subsequent feature selection procedures, allowing for integrating various representations and information sources within the model. By infusing prior knowledge in this manner, LFIT capitalizes on the insights provided by qualitative descriptions of the surrounding conditions for each monitoring point. This integration enhances the model's ability to capture critical contextual contents and their potential influence on landslide behavior.

\subsubsection{ Step 2:  Using dynamical feature selection to identify and reveal driver of landslide behavior across scales and periods}
\label{subsection 3:3:2  Using dynamical feature selection to identify and reveal driver of landslide behavior across scales and periods}

The key to employing dynamic feature selection for identifying and revealing drivers of landslide behavior across scales and periods in LFIT is GRN units. The gate mechanism of GRN enables capturing intricate feature interactions influencing landslides. Specifically, high-level representation extraction is substantially enhanced through parameterized projections $\boldsymbol{W}(\cdot) \in \mathbb{R}^{d_m \times d_m}$ and $\boldsymbol{b}(\cdot) \in \mathbb{R}^{d_m}$. Input features most informative at each spatiotemporal point become more prominent when the neural network's attention centers on them. By detecting and utilizing these subtle yet crucial patterns within the inputs, we can determine how landslides deform and capture the changing impacts of covariate factors on landslide deformation states. The selective focus ability of GRN is well suited to the complexity and diversity inherent in landslide knowledge and data for grasping what governs landslide dynamics in all their variety.

GRN operates on the primary input $\boldsymbol{a}$ and optionally the context input $\boldsymbol{c}$. The input data refer to high-dimensional information rather than raw measurement data, which have undergone mapping before GRN input. Layer by layer, landslide spatiotemporal dynamics can be efficiently encoded by progressively extracting increasingly abstract and informative representations $\gamma_1$. The operation of a basic GRU unit can be described as follows.

$$
\begin{aligned}
\operatorname{GRN}(\boldsymbol{a}, \boldsymbol{c}) & =\text { LayerNorm }\left(\boldsymbol{a}+\operatorname{GLU}_\omega\left(\boldsymbol{\gamma}_1\right)\right) \\
\boldsymbol{\gamma}_1 & =\boldsymbol{W}_{1, \omega} \boldsymbol{\gamma}_2+\boldsymbol{b}_{1, \omega} \\
\boldsymbol{\gamma}_2 & =\operatorname{ELU}\left(\boldsymbol{W}_{2, \omega} \boldsymbol{a}+\boldsymbol{W}_{3, \omega} \boldsymbol{c}+\boldsymbol{b}_{2, \omega}\right)
\end{aligned}
$$

One of the most critical components in the LFIT architecture, i.e., Variable Selection (VS) blocks, are constructed upon distinct functional GRU for handling all input vectors. The VS blocks comprise two parts. In the first part, $\mathrm{GRN}_\alpha$ independently filters each input vector, differentiating informative features from noisy information. The filtered vectors are obtained as $\tilde{z}_t^{(j)}=\operatorname{GRN}_\alpha\left(\boldsymbol{z}_t^{(j)}\right)$. The second part involves attention-like operations. Here, $\mathrm{G R N}_\beta$ accepts the flattened inputs $\mathcal{Z}_t=\left[\boldsymbol{z}_t^{(1)^T}, \ldots, \boldsymbol{z}_t^{\left(n_X\right)^T}\right]^T$, alongside additional input variables represented by $\boldsymbol{c}_v$, which captures the prior knowledge. Softmax activation is utilized to assign weights to the output of $\mathrm{G R N}_\beta$, enabling each input variable to contribute to the landslide displacement prediction targets with a specific degree. The weight vector $\mathcal{W}_v$ is calculated as $\mathcal{W}_v=\operatorname{Softmax}\left(\operatorname{GRN}_\beta\left(\mathcal{Z}_t, \boldsymbol{c}_v\right)\right)$. The resulting output represents an automatic selection of meaningful information for landslide displacement forecasting, expressed as $\tilde{\boldsymbol{z}}_t=\sum_{j=1}^{n_{X}} \mathcal{W}_v^{(j)} \tilde{\boldsymbol{z}}_t^{(j)}$, which denotes the linear combination of weights and each filtered input variable.

\begin{figure*}[!ht]
	\centering
	\captionsetup{labelfont=bf}
	\includegraphics[width=16.5cm]{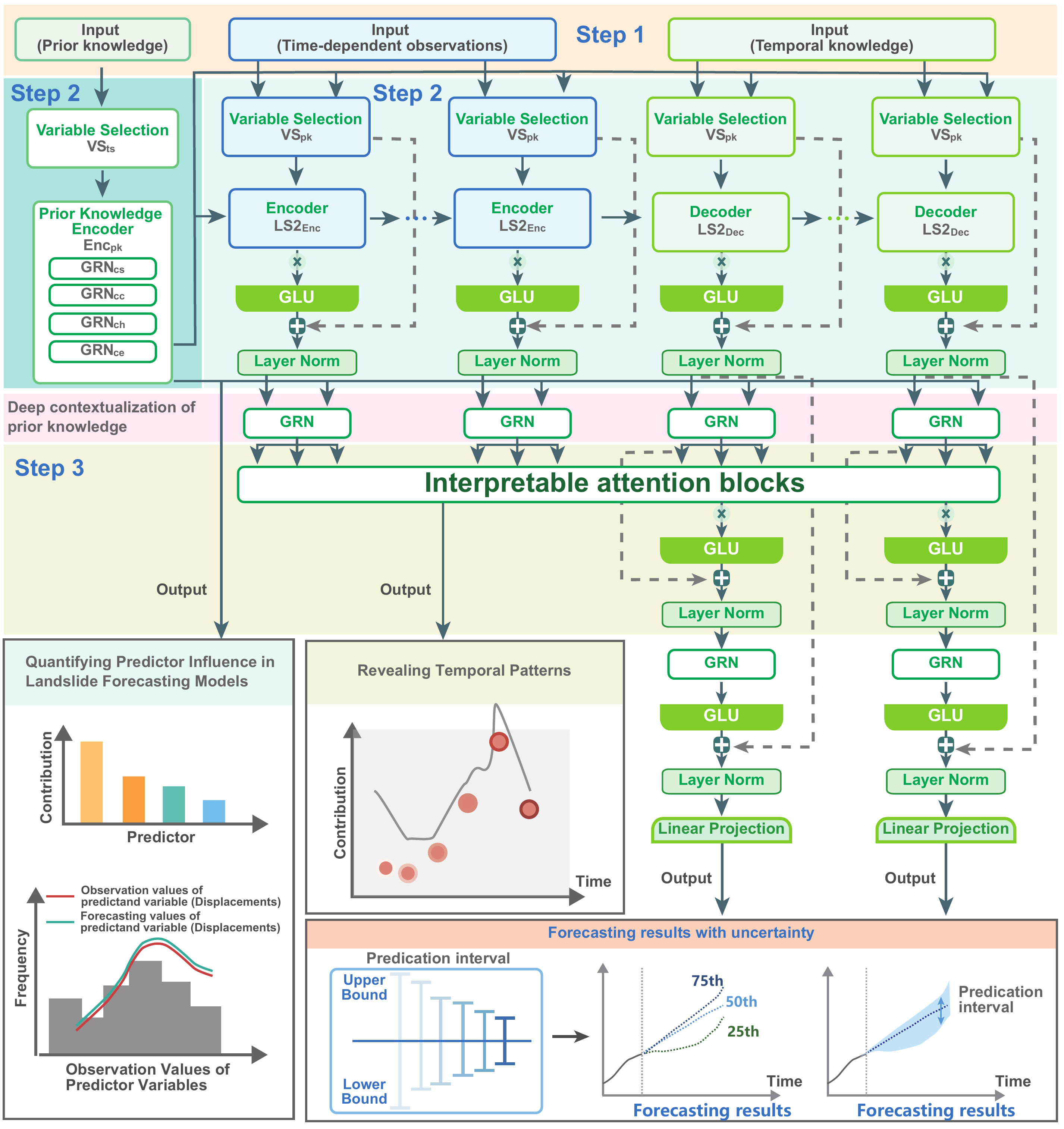}
	\caption{\textbf{The overall framework for LFIT involves three successive steps.} \textbf{Step 1:} Embeddings, wherein Prior Knowledge and Data are projected into high-dimensional space. \textbf{Step 2:} Dynamic identification and revelation of landslide behavior drivers across various scales and periods. \textbf{Step 3:} Identification and uncovering of temporal response patterns to influencing factors across spatial regions of the landslide. Additionally, the framework incorporates an optimization process that accounts for data uncertainty, ultimately yielding forecasting results with prediction intervals.} 
	\label{Figure2}
\end{figure*}

As illustrated in Fig. \ref{Figure2}, there are distinct differences between the Variable Selection (VS) blocks used for selecting time-dependent features ($\mathrm{VS}_{ts}$)and prior knowledge features ($\mathrm{VS}_{pk}$), primarily stemming from the incorporation of optional external context information. Notably, the external context $\boldsymbol{c}_v$ refers to information extracted by the Prior knowledge encoder ($\mathrm{Enc}_{pk}$). Since $\mathrm{VS}_{pk}$ inherently encompasses prior knowledge, it does not necessitate any additional prior knowledge information for feature selection guidance.

$\mathrm{Enc}_{pk}$, also known as static covariates encoding, assumes a crucial role in enabling neural networks to effectively analyze the intricate dynamics of landslide deformation processes. It establishes a connection between local landslide properties, the temporal behavior of landslide displacement, and the variations in the external environment. To propagate this prior information, four GRN modules are employed to extract meaningful representations from the prior knowledge. Each GRN module provides a denoised prior knowledge representation that can be applied at different stages of the architecture, gradually guiding the neural network's learning process for landslide displacement under complex factors. In layer 2, $\mathrm{GRN}_{cs}$ utilize prior knowledge representations to guide $\mathrm{VS}_{ts}$ in obtaining more relevant features. These features encompass information that captures both time-dependent measurements and descriptive prior knowledge pertaining to local conditions, allowing for a comprehensive understanding of hidden impacts across multiple influencing factors. Furthermore, a $\mathrm{GRN}_{cc}$ in layer 3 and a $\mathrm{GRN}_{ch}$ in layer 4 contribute prior knowledge representations that serve as the initialization for the cell state and hidden state of the LSTM layers, respectively. This strategic integration guides the recurrent neural networks to uncover local temporal patterns of landslide displacement in the presence of complex interactions among various influencing factors. Finally, in layer 5, the $\mathrm{GRN}_{ce}$ module provides representations that further integrate the prior knowledge with other time-dependent data. This integration enhances the LFIT's ability to capture the comprehensive effects of prior knowledge in conjunction with other relevant factors, ultimately contributing to more accurate and insightful forecasting of landslide deformation behavior.

The static representations from $\mathrm{Enc}_{pk}$ capture prior knowledge about landslide properties in the local monitoring area. However, directly blending these categorical features with time-dependent data would introduce noise and deteriorate forecasts. Therefore, the model should disentangle static prior knowledge from time-varying data so that the static features can guide the understanding of temporal patterns without disrupting the learned dynamics. LFIT employs LSTM-based sequence-to-sequence (LS2) layers (Layers 9-10) to decouple static and time-varying representations. Specifically, the prior knowledge embeddings $\boldsymbol{c}_h$ and $\boldsymbol{c}_c$ from $\mathrm{Enc}_{pk}$ are used to initialize the LSTM hidden state and cell state, respectively. This conditions the LSTM context embeddings on prior knowledge while maintaining the learned temporal dynamics. The hidden state ht and cell state ct of the LSTMs can thus be expressed as:

$$
\begin{aligned}
& h_t=\mathcal{F}\left(x_t, h_{t-1}, \boldsymbol{c}_h\right) \\
& c_t=\mathcal{G}\left(x_t, c_{t-1}, \boldsymbol{c}_c\right)
\end{aligned}
$$
where $x_t$ is the input at time $t$, and $\mathcal{F}$ and $\mathcal{G}$ are the LSTM transition functions for the hidden state and cell state, respectively.

By initializing the internal states of the LSTMs with prior knowledge, the model can relate the temporal behavior of landslide displacement to static landslide properties and surrounding conditions. The LSTM context embeddings capture the dynamics of landslide displacement while being appropriately conditioned on prior knowledge.

\subsubsection{ Step 3:  Revealing Temporal Patterns of Response to Influencing Factors across Landslide Spatial Regions}
\label{subsection 3:3:3  Revealing Temporal Patterns of Response to Influencing Factors across Landslide Spatial Regions}

LSTM-based sequence-to-sequence models (LS2) are used to model the temporal dynamics of landslide behavior (here, deformation measurements) at each monitoring point (Layers 11-12). LS2 can identify significant time points like anomalies, change points, and cycles by learning local connections between temporally proximate states.

The encoder ($\mathrm{LS2}_{Enc}$) processes static inputs and time-dynamic future-known inputs $\tilde{\mathcal{Z}}_{t-k: t}$ before passing them on to the decoder ($\mathrm{LS2}_{Dec}$). $\mathrm{LS2}_{Dec}$ incorporates inputs that are unknown in the future. At the LS2 phase, the result is $\iota(t, n) \in\{\iota(t,-k), \ldots, \iota(t, \tau)\}$, which can be viewed as an abstract representation of how multiple factors interact with each other and affect landslide displacements. $n$ enable its explicitly having temporal meaning, which denotes an index of sequential positions to improve understanding of temporal order in subsequent self-attentive layers. Further shallow feature selection involves gate operation to remove noise and residual connection to add possible missing information, which results in final information extraction $\tilde{\imath}(t, n)$. During layer 12, further nonlinear processing enhances the guide given by prior knowledge. During this phase, the priori representation provided by $\mathrm{GRN}_{ce}$ above informs other temporal representations about specific characteristics of regions where monitoring points are located. $\mathrm{GRN}_{ch}$ are blocks responsible for implementing such non-linear transformations. $\mathrm{GRN}_{eh}$ further processes the resulting temporal representation $\tilde{\imath}(t, n)$ derived from LS2 and $\boldsymbol{c}_e$ from previous $\mathrm{GRN}_{ce}$, and obtain final feature extraction result by $\zeta(t, n)=\operatorname{GRN}_{e h}\left(\tilde{t}(t, n), \boldsymbol{c}_e\right)$.

LSTM inherently acts as positional encoding, enabling each state in the high-level representation of each layer to correspond to information at a specific time point. At the high-level representation of each layer, each state represents information at a particular time point, and states are linked in a sequence to represent a temporal progression. Therefore, LSTM effectively preserves the sequential nature of the time series data, ensuring that the temporal dependencies are maintained. This explicit connection between learned representations and specific time points is important for understanding landslide deformation patterns, where long timescales and complex interactions between factors are involved. Each state encapsulates the intricate and evolving dynamics resulting from the interplay of multiple factors and the corresponding displacement response over time. By explicitly encoding the temporal location within the learned representations, LFIT gains the ability to discern the dynamically changing influences of various factors and their corresponding responses over time.

Layer 13 employs a self-attentional mechanism to gain insights into landslide forecasts. Self-attention attends to all time steps in the learned holistic representation, capturing short- and long-term dependencies across the entire series. Formally, for a series $x \in \mathbb{R}^{\mathcal{T}}$ of length $\mathcal{T}$, projection matrices $\mathbf{W}_Q, \mathbf{W}_K, \mathbf{W}_V \in \mathbb{R}^{d \times d}$ transform $x$ into query $\mathbf{Q}$, key $\mathbf{K}$, and value $\mathbf{V}$ matrices. Scaled dot-product attention can be obtained by $\mathcal{A}(\mathbf{Q}, \mathbf{K})=\operatorname{Softmax}\left(\frac{\mathrm{QK}^T}{\sqrt{d}}\right)$. Thus, the self-attention scores can be calculated by $\mathcal{A}t t(\mathbf{Q}, \mathbf{K}, \mathbf{V})=\mathcal{A}(\mathbf{Q}, \mathbf{K}) \mathbf{V}$. Self-attention with multiple heads is able to focus a single input on different subspaces, allowing for more nuanced attention, which in turn allows for the extraction of subtle features and the capture of diverse behaviors observed over long time scales in landslide deformation. 

As with conventional multiheaded self-attention, interpretable attention blocks in LFIT learn distinct temporal patterns from each head. Different from other attention blocks, the attention block allows each head to be aware of different temporal patterns while attending to the same input. Attention weights are combined into matrices that can be visualized and interpreted. This attention mechanism facilitates the exploration of relationships between input elements and the interpretation of global features. It does this by emphasizing significant landslide states at specific time points and their interactions with other information. By concatenating the $h$-th extracted features for input variables, temporal attention $\tilde{\mathbf{F}}_{t a}$ can be obtained, which is a representation of the temporal dynamics, that can include multiple temporal patterns in longer time series. The process can be described as follows.

$$
\begin{aligned}
\tilde{\mathbf{F}}_{t a} & =\tilde{\mathcal{A}}(\mathbf{Q}, \mathbf{K}) \mathbf{V} \mathbf{W}_V \\
& =\left\{1 / \mathcal{H} \sum_{h=1}^{\mathcal{H}} A\left(\mathbf{Q} \mathbf{W}_Q^{(h)}, \mathbf{K} \mathbf{W}_K^{(h)}\right)\right\} \mathbf{V} \mathbf{W}_V \\
& =1 / \mathcal{H} \sum_{h=1}^{\mathcal{H}} \mathcal{A}t t\left(\mathbf{Q} \mathbf{W}_Q^{(h)}, \mathbf{K} \mathbf{W}_K^{(h)}, \mathbf{V} \mathbf{W}_V\right)
\end{aligned}
$$

Further insights emerge from analyzing attention score outputs. Attention weight patterns reveal the most significant past displacement states for predictions at each time step. $\tilde{\mathcal{A}}(\mathbf{Q}, \mathbf{K})$ gives a score quantifying how much a past state impacts future state prediction. Scores identify states that most influence predictions, uncovering patterns the model autonomously learns without predefined trend or seasonality concepts. This attention mechanism also serves to quantify each input's contribution to forecasts at specific time steps by extending its temporal attention to other related inputs.

Furthermore, feature selection blocks can also be utilized to better understand the drivers behind LFIT's predictions and their dynamics. In feature selection blocks, all inputs are weighted with a softmax function, which gives emphasis to relevant features. We gain insight into which inputs play a significant role in determining LFIT's behavior. Formally, the inputs covering various information relating to landslide displacement are processed by a series of non-linear transformation in neural networks and are denoted as $\mathbf{F}_{g r}$. Formally, the inputs covering various information relating to landslide displacement are processed by a series of non-linear transformation in neural networks and are denoted as $\mathbf{F}_{g r}$ and they are weighted using the softmax function to obtain the weighted representation with d hidden dimensions. The process can be expressed by $\mathbf{F}_{f a}=\operatorname{Softmax}\left(\mathbf{F}_{g r}\right)$. Weight matrix $\mathbf{F}_{f a}$ is utilized to identify feature contributions. All elements add up to one, each corresponding to a separate time step and a different information channel.

\subsection{ Training with Optimization for Handling Data Uncertainty}
\label{subsec:2:4:Training with Optimization for Handling Data Uncertainty}

Here, we employ a quantile regression loss function, which can estimate arbitrary quantiles of the response variable, not just the mean, providing information on variability and uncertainty. By incorporating the quantile loss function into the LFIT framework, we construct prediction intervals that serve as plausible bounds, accounting for the aleatoric uncertainty inherent in future landslide dynamics. Consequently, our approach yields a more comprehensive and reliable assessment of potential future variations in landslide deformation and other behavior-related measurements.

This loss function can be calculated by $\mathbf{Q S}(\tilde{y}, y, q)=\max \left(\left\{q\left(y_i-\tilde{y}_i\right),(1-q)\left(y_i-\tilde{y}_i\right)\right\}\right)$, where $\tilde{y}_i$ is the the estimated quantile, $y_i$ is the target value, $q \in[0,1]$ is the quantile of interest. The larger the value of q, the greater the penalty for over-prediction. For global forecasting of landslide behavior involving multiple variables, we optimize an objective function ($\left(\mathcal{L}_{\text {LFIT }}\right)$) that incorporates all quantiles and the forecasting targets over the prediction period, thus providing a probabilistic perspective on how the landslide system might evolve over time under different conditions.

$$
\mathcal{L}_{\mathrm{LFIT}}(\Omega, \Theta)=\sum_{y_t \in \Omega} \sum_{q \in \mathcal{Q}} \sum_{\tau=1}^{\tau_{\max }} \frac{\mathbf{Q S}(\tilde{y}(q, t-\tau, \tau), y, q)}{\mathcal{B} \tau_{\max }}
$$

The objective function ($\left(\mathcal{L}_{\text {LFIT }}\right)$) is optimized via backpropagation using mini-batch gradient descent to determine the parameters of the neural network ($\Theta$). It incorporates the quantile loss for the variables, and the forecasting targets ($\Omega$), including $\mathcal{B}$ observations. $\tau_{\max }$ represents the farthest forecasting time point. $Q$ is the set of quantiles to be estimated, which here is [0.2,0.1,0.25,0.5,0.75,0.9,0.98].

\section{Results}
\label{sec:3: Results}

We present a knowledge-guided landslide forecasting pipeline, employing an interpretable transformer-based neural network, LFIT. We emphasize three primary components regarding this architecture:

\textbf{1. Gate mechanisms for data and knowledge fusion:} Gate mechanisms enable the integration and focus of key information regarding landslide behavior, allowing the model to retain and utilize essential information for accurate forecasting while filtering out extraneous information. By incorporating attention-like mechanisms and basic gate mechanism blocks at varying scales, variable selection network blocks can automatically adjust attention weights, prioritizing input features based on their relevance, thus eliminating the manual feature selection process and providing insights into the drivers behind forecasting results and underlying landslide dynamics.

\textbf{2. LSTM-based encoders and decoders for temporal dynamics preservation: } LFIT employs LSTM networks within its encoder-decoder architecture to effectively model the complex spatiotemporal dynamics of landslide monitoring data. The LSTM encoders can capture intricate relationships across multiple time steps and disentangle the representations of different predisposing factors (e.g., rainfall, and soil moisture) that drive landslide occurrence. By leveraging LSTMs' ability to retain long-term dependencies, LFIT develops a comprehensive understanding of the temporal patterns underlying landslide deformation and forecasting. The attention mechanism within the LSTMs assigns weights to each time step, enabling the model to focus on the portions of the input history that are most relevant for predicting local landslide dynamics. These attention weights signify the model's perception of how significantly different predisposing factors (e.g., rainfall, soil moisture) at each time step influence landslide occurrence at a given location. By visualizing how the attention weights change over space and time, we can gain insights into the key predisposing factors influencing different parts of landslides.

\textbf{3. Quantile regression loss function for reliable forecasting: }To ensure dependable and effective forecasting results, LFIT incorporates a quantile regression loss function to construct prediction intervals that account for uncertainty in predictions. This approach minimizes and quantifies aleatoric uncertainty arising from noise rising from the input.
Detailed information about these components can be found in the method section and the supplement table.

An AMD Ryzen 9 5900HX processor, a Nvidia GeForce RTX 3080 graphics card, along with 32 GB of DDRAM, were used in the experiment. The implementation utilized PyTorch 1.10, based on Python 3.9 in the Anaconda distribution. We utilized PyTorch Forecasting, a software package that implements state-of-the-art deep learning methods for time series forecasting. A more automated approach to data preprocessing is offered by this package, simplifying the process and improving forecasting efficiency. 

Network training follows principled practices to ensure reproducible results.The dataset was initially partitioned into training and testing sets. Hyperparameters were chosen based on prior research to minimize overfitting risk, without additional tuning. Our study did not attempt to fully explore deep learning models through extensive hyperparameter fine-tuning, but it should be noted that minor adjustments can be made experimentally to optimize performance, taking dataset size into account. We implemented modifications to enhance model performance, including dropout-enabled layers and applying early stopping techniques.
The training procedure also encompasses the following aspects: (1) A batch size of 128 balances computational efficiency with model convergence. (2) Standardization is applied to maintain consistent data representation by subtracting the mean and scaling the data to unit variance. (3) The Adam stochastic gradient descent optimizer is chosen as the default option, featuring a weight decay of 5 to encourage regularization. (4) Early stopping enhances the model's generalization capacity and mitigates overfitting, ensuring robust and reliable performance on previously unseen data.

\subsection{Exploring the Influence of Prior Knowledge and Environmental Variables on Landslide Forecasting: Experimental Scenarios}
\label{subsec:3:1:Exploring the Influence of Prior Knowledge and Environmental Variables on Landslide Forecasting: Experimental Scenarios}

Here, we propose a pipeline that seamlessly integrates domain knowledge into a data-driven framework. Our primary focus is on incorporating relevant prior information from historical documents and expert insights into the geological conditions of each monitoring site, including qualitative descriptions of geotechnical properties. Our hypothesis posits that observations of landslide behavior at individual monitoring sites reveal distinct geological influences. In this context, a displacement time series for a specific observation point captures the deformation characteristics of a distinct patch within the landslide system. Then, by incorporating categorical data as prior knowledge, our approach can capture semantic differences across diverse locations on the landslide body. In this regard, monitoring point labels act as descriptive descriptors of their corresponding patches, thus supplementing prior knowledge inputs. 

Our comprehensive pipeline integrates environmental data from multiple sources, including meteorological measurements and landslide behavior metrics, to effectively investigate the complex spatiotemporal dynamics of landslides influenced by heterogeneous geological conditions and site-specific triggering mechanisms. A wide range of inputs we considered include weather factors (precipitation, surface temperature, surface humidity, and reservoir water level fluctuations), prior qualitative knowledge from domain experts (regional hazard level variations and geotechnical properties), and landslide characteristics (e.g., elevation).

\begin{figure*}[!ht]
	\centering
	\captionsetup{labelfont=bf}
	\includegraphics[width=16.5cm]{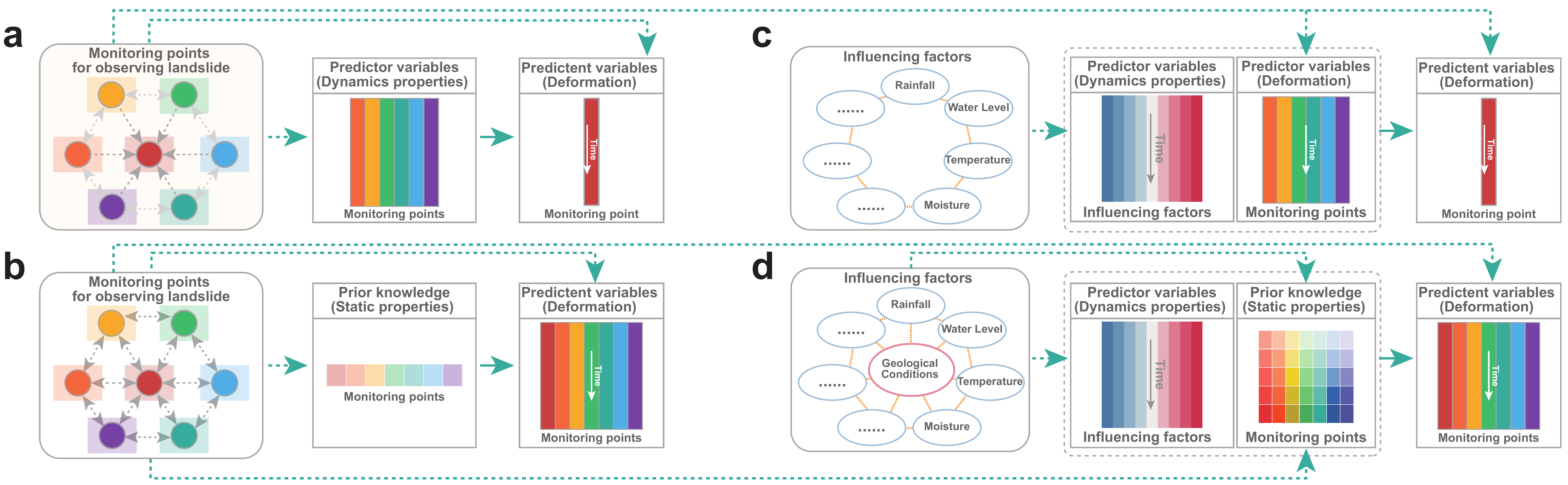}
	\caption{\textbf{Experimental scenarios} \textbf{a.} Single-Target forecasting with Neighboring Spatial Measurements from Surrounding Monitoring Points (\textbf{ST-NSP}).  \textbf{b.} Multi-Target Forecasting with Monitoring Point Categorical Information (\textbf{MT-MPC}). \textbf{c.} Single-Target forecasting with Environmental Variables (\textbf{ST-NSP-EV}). \textbf{d.} Multi-Target Forecasting with Integrated Prior Knowledge, Environmental Variables (\textbf{MT-MPC-PK-EV}).} 
	\label{Figure3}
\end{figure*}

As shown in Fig.\ref{Figure3}, we designed and executed four experimental scenarios to evaluate the effectiveness of incorporating diverse information types in knowledge-infused landslide forecasting using the interpretable neural network LTFI. 

Scenario 1 (\textbf{ST-NSP}): Single-Target forecasting with Neighboring Spatial Measurements from Surrounding Monitoring Points. This scenario focused on forecasting a single target using monitoring data from surrounding points. Based solely on neighboring measurements, we perform interpretable landslide forecasting without considering prior knowledge or environmental factors.

Scenario 2 (\textbf{MT-MPC}): Multi-Target Forecasting with Monitoring Point Categorical Information. This scenario involves multiple variable forecasting across multiple landslide monitoring sites. We integrate monitoring point information as categorical prior knowledge, leveraging each monitoring site's interdependencies and unique characteristics to predict interdependent time series at various locations. This scenario emphasizes location-specific information' role in reflecting landslide conditions without incorporating environmental factors. This allows us to illustrate the importance of monitoring points as categorical information for capturing and understanding the dynamics of the landslide system.

Scenario 3 (\textbf{ST-NSP-EV}): Single-Target forecasting with Environmental Variables. In the third scenario, we consider the same landslide behavior-related measurements and relevant environmental variables nearby. By incorporating these external factors, we aimed to capture the broader context and potential influences on the target variable.

Scenario 4 (\textbf{MT-MPC-PK-EV}): Multi-Target Forecasting with Integrated Prior Knowledge, Environmental Variables. In the fourth scenario, we expand the scope of location-specific spatial information by including additional categorical variables. Alongside monitoring point labels, we encode qualitative descriptions of geotechnical properties and hazard levels as categorical variables. This prior spatial knowledge gives context to the monitoring points. Additionally, environmental factors like climate and topography were incorporated to capture a broader range of predictive features and provide insights into the combined effects of various predictors.

These experimental scenarios allowed the systematic examination of the impact of various data inputs and prior knowledge on the understanding and predictability of landslide behavior and the drivers that drive landslides. We conducted our evaluation by gradually increasing the complexity of the input data, beginning with merely surrounding monitoring data and successively incorporating monitoring data, prior knowledge, additional attributes, and environmental variables. Evaluation metrics, such as Mean Absolute Error (MAE), Mean Absolute Percentage Error (MAPE), Root Mean Squared Error (RMSE), and Symmetric Mean Absolute Percentage Error (SMAPE), were employed to quantify the predictive accuracy of landslide forecasting. Further details regarding these evaluation methods can be found in the Supplementary Material.

\subsection{Forecasting deformation and Deciphering Influencing Factors for Reservoir Landslide}
\label{subsec:3:2:Forecasting deformation and Deciphering Influencing Factors for Reservoir Landslide}

Reservoir landslides, characterized by complex thermo-hydro-mechanical interactions, present a significant risk to the safety of dams, infrastructure, and communities near reservoir banks. The growing frequency and intensity of extreme climate events, encompassing temperature fluctuations, precipitation variations, and flooding, have increased the vulnerability of infrastructure networks in reservoir regions. In recent years, this heightened vulnerability has led to catastrophic landslides with cascading effects. A prime example is the Three Gorges Reservoir (TGR) in China, where unprecedented rainfall exceeding 500 mm occurred between May and July 2020, resulting in the most severe flood year since 1998. The extreme weather conditions caused numerous landslides to move and collapse, highlighting the need for more advanced forecasting technologies to mitigate risks.

Here, we examine the influence of infusing knowledge into deep learning architectures on the predictability of reservoir-triggered landslides and aim to address several key questions. The first question is: How does the infusion of knowledge into deep learning architectures affect spatial and temporal patterns within LFIT? Can we enhance landslide forecasting accuracy and interpretability by incorporating relevant information and prior knowledge? Second, can interpretable landslide deformation using LTFL capture physically plausible patterns and reveal meaningful associations with influencing factors? Finally, can the insights gained from the LTFL approach be generalized to other regions facing similar challenges?

\begin{figure*}[!ht]
	\centering
	\captionsetup{labelfont=bf}
	\includegraphics[width=16.5cm]{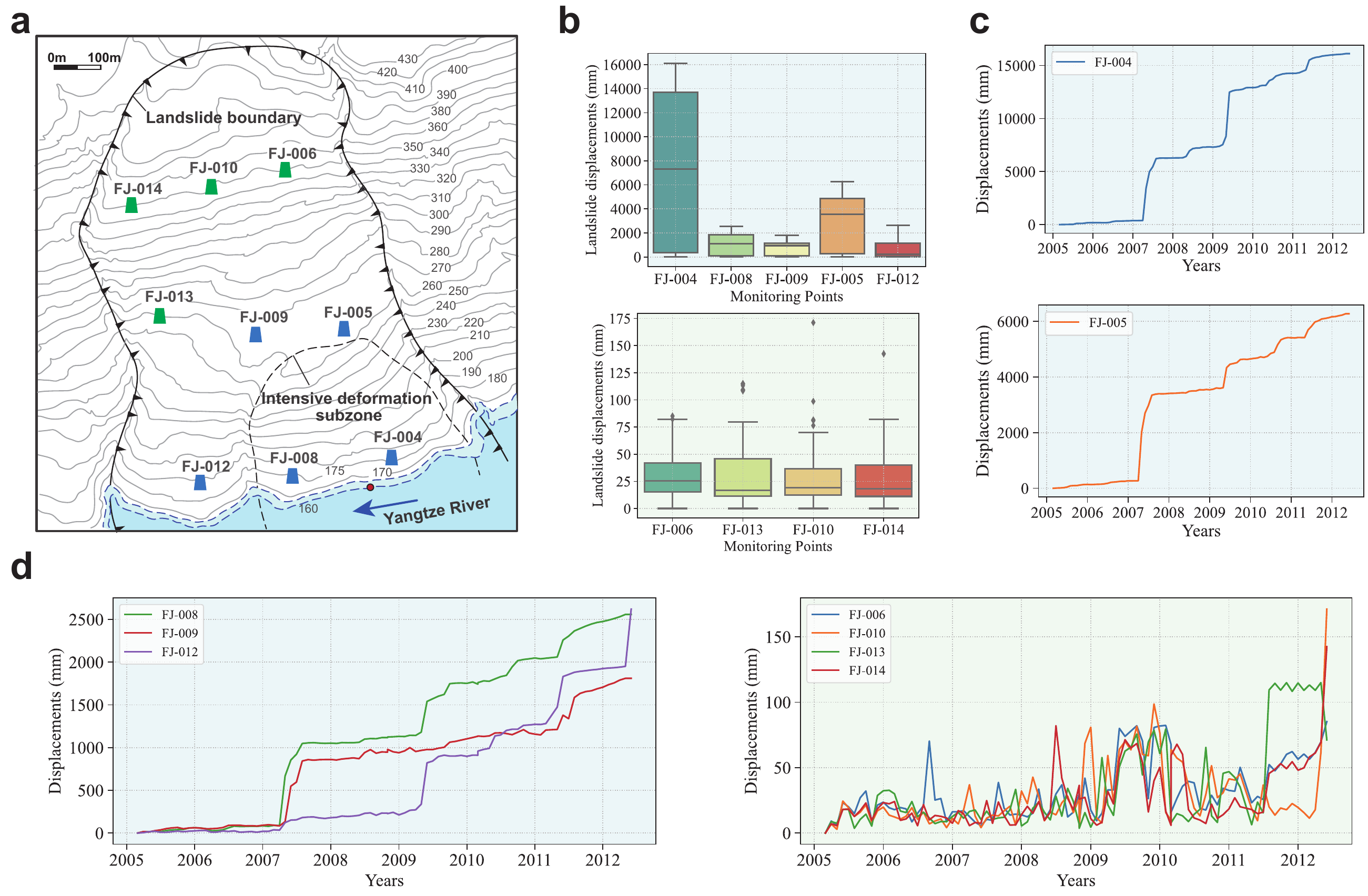}
	\caption{\textbf{Reservoir Landslide Cases} \textbf{a.} Observation point in reservoir landslide.  \textbf{b-d.} Characteristics of deformation at different observation points. At the green observation points, deformation measurements show similar characteristics, including small deformations, oscillatory displacement curves, and significant increases prior to landslide failure. By contrast, measurements at the blue observation points show distinct characteristics, including step-like displacement curves and more significant deformations.} 
	\label{Figure4}
\end{figure*}

Our study case is the Huanglianshu landslide in Three Gorges Reservoir, a reservoir-triggered landslide monitored since 2003. Deformation measurements indicate that the observed deformation pattern varies significantly between the various areas of the site monitored. Fig.\ref{Figure4} illustrates the contrasting deformation patterns observed at different monitoring points. At the leading edge is a substantial amount of deformation, characterized by step-like displacements with distinct turning points and periodic characteristics between significant displacements. As a comparison, the deformation pattern at the middle and rear edges exhibits smaller magnitudes, with fluctuations exhibiting abrupt changes in magnitude. A detailed description of the dataset can be found in the Methods section.

We trained LTFL to forecast the monthly deformation of this landslide, considering various factors known to influence landslides as inputs. For example, monthly records of precipitation and water level fluctuations in reservoirs are typically critical triggers for such landslides. Furthermore, we incorporated categorical information about each landslide time series, providing expertise-based prior knowledge. The monitoring points are each associated with essential information, such as the danger level and the geotechnical characteristics of the area. We incorporated these categorical labels to enrich our understanding of each landslide time series's unique characteristics and behaviors. Additionally, we utilized temporal labels, such as year, month, and day, as supplementary prior knowledge. These temporal labels were crucial in capturing the underlying temporal structure within the data, enabling the identification of patterns and trends associated with specific periods, such as seasonal effects and annual trends.

Our analysis investigates three interconnected aspects, emphasizing the interpretability and physical consistency of landslide behavior as captured by data from landslide monitoring systems and their impact on landslide forecasting performance. We visualize the acquired knowledge, such as non-adjacent relationships across multiple observational sites and temporal patterns, thus elucidating the underlying natural laws and physical mechanisms governing landslides, including spatial connections and lagged correlations, thereby obtaining insights into the complex spatiotemporal patterns observed during landslide forecasting:(1) We examine spatial awareness in the aforementioned four experimental scenarios, evaluating the implications of integrating contextual information on the dynamics of spatial patterns.(2) We investigate temporal awareness by examining variations in spatial and temporal scales, assessing the changes in temporal awareness's impact on forecasting accuracy, and understanding the trigger mechanisms according to the learned temporal patterns.
(3) We analyze the influencing factors driving the improved prediction results to discern the fundamental drivers governing the accuracy and reliability of the forecasting outcomes.

\subsubsection{Interpretable Landslide Forecasting: Enhancing Neural Networks Spatial Awareness}
\label{Interpretable Landslide Forecasting: Enhancing Neural Networks Spatial Awareness}

We first evaluated the variability of the forecasting performance achieved in different landslide parts and four experimental scenarios using the results of RMSE calculations. We analyzed five monitoring points on the leading edge and four on the middle and trailing edges, respectively. 

\begin{figure*}[!ht]
	\centering
	\captionsetup{labelfont=bf}
	\includegraphics[width=16.5cm]{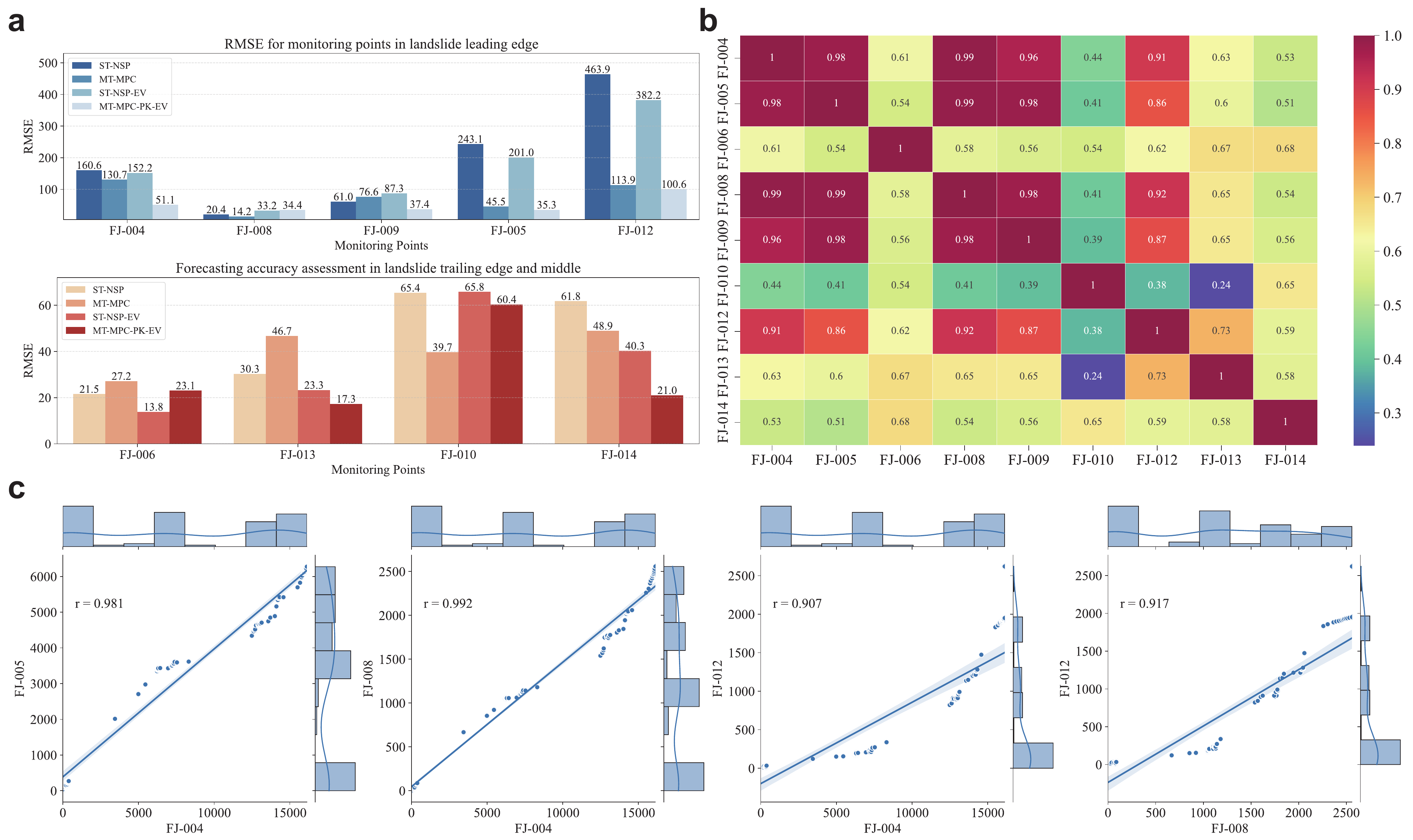}
	\caption{\textbf{Forecasting Results and Spatial Analysis across Multiple Observation Points} \textbf{a.} Comparative Evaluation of Forecasting Results using Metrics.  \textbf{b-c.} Quantifying Correlations among Deformation Observations at Monitoring Sites.} 
	\label{Figure5}
\end{figure*}

Results reveal that the best forecasting performance is observed at the leading edge of the landslide, although the deformation is the largest (Fig.\ref{Figure5}). In this region, considering the monitoring point names as a priori category information and conducting simultaneous forecasts for multiple points (Scenario \textbf{MPC}) yields the lowest root mean square error (RMSE). However, it is essential to note that this pattern does not consistently apply to the middle and trailing edges, as exemplified by points FJ-006 and FJ-013. The performance in these areas is further enhanced by incorporating additional information on driving factors (Scenario \textbf{EV}) and geotechnical properties (Scenario \textbf{PK-EV}). Notably, the most comprehensive scenario, which combines multi-source data integration and a priori knowledge (\textbf{MT-MPC-PK-EV}), achieves the highest forecasting accuracy.

Fig.\ref{Figure5}b illustrates the Pearson correlation coefficients among multiple monitoring points, revealing synchronized deformation patterns and indicating a consistent response in certain landslide parts to external factors due to shared mechanisms. Notably, monitoring points at the leading edge demonstrate a strong correlation, thus highlighting the interconnected nature of landslide deformation dynamics. This correlation significantly enhances the accuracy of our holistic forecasting approach and emphasizes the importance of considering spatiotemporal relationships among monitoring points to capture the interconnected dynamics of landslide behavior. Furthermore, we have highlighted the importance of considering spatial relationships and interdependencies between monitoring points to forecast landslide deformation. By integrating diverse data sources and leveraging prior knowledge, we enhance spatial awareness, leading to improved forecasting performance in regions with stronger correlations.

\begin{figure*}[!ht]
	\centering
	\captionsetup{labelfont=bf}
	\includegraphics[width=16.5cm]{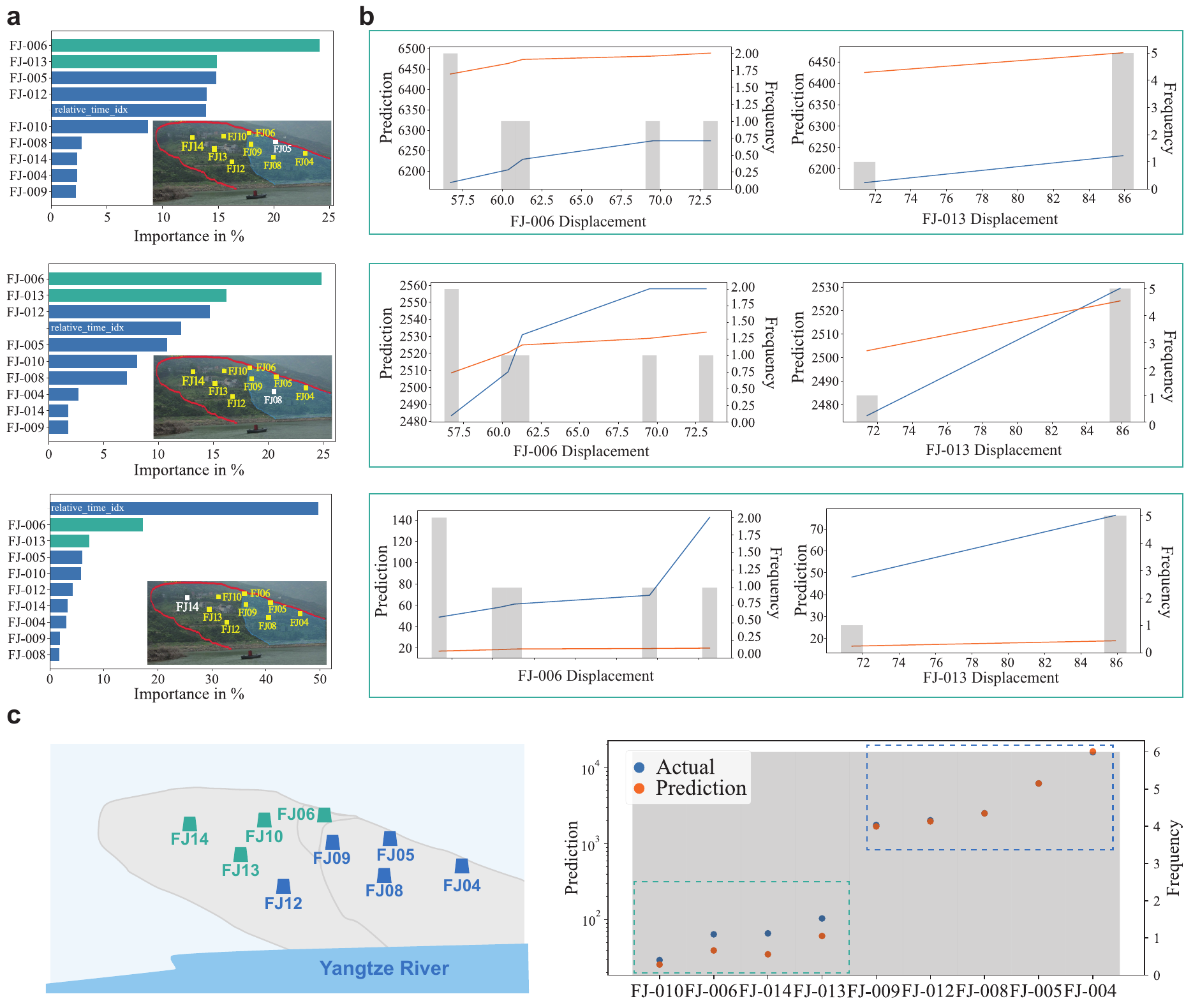}
	\caption{\textbf{Spatial Awareness and Interpretability of LFIT Model Performance and Robustness in Landslide Deformation Forecasting.} \textbf{a.} Assessment of monitoring point variables' impact on prediction outcomes in the \textbf{ST-NSP} scenario, detecting rare events and model limitations.   \textbf{b.} Depiction of the \textbf{ST-NSP} scenario characterized by increased error metrics and significant forecasting inconsistencies. \textbf{c.} Multi-target forecasting scenario (\textbf{MT-MPC}) illustrating the prediction accuracy for deformation measurements at location-specific monitoring points. The degree of overlap between blue and orange dots indicates the forecasting accuracy, highlighting challenges posed by oscillatory patterns and abrupt alterations in deformation data.} 
	\label{Figure6}
\end{figure*}

The interpretability provided by LFIT allows for a further understanding of spatial correlations in our four scenarios, each considering spatial relationships in distinct ways. In the ST scenario, the deformation data from other monitoring points are considered covariates. The feature selection blocks of LFIT quantitatively assess the importance of each input variable, facilitating the estimation of contributions to single target forecasting from neighboring monitoring points' deformation measurements. As a result, the feature selection module provides the deformation contribution from other monitoring points to the current prediction target. Fig.\ref{Figure6}a illustrates the forecasting results for the \textbf{ST-NSP} scenario, highlighting the influence of displacement time series from other monitoring points on the prediction outcomes for FJ-005, FJ-008, FJ-013, and FJ-014. In the displacement predictions for monitoring points FJ-004 and FJ-008, the deformation series from other monitoring points demonstrate comparable importance, implying similar influences on these targets.

Furthermore, the interpretable nature of LFIT allows for analyzing the robustness of other variables within their value ranges. Capturing abrupt changes in landslide deformation becomes crucial when multiple landslide displacement variables are used as covariates. Often elusive and mistaken for random noise, these rare events challenge the model's ability to capture their behavior. To address this, we plot the fluctuations in the predicted shape of the monitoring points that exert the greatest influence on the target prediction. The gray bars in the plot represent the distribution of each variable, highlighting the frequencies of different values.  The agreement between the blue and orange curves at low frequencies determines the model's ability to capture rare events, such as sudden changes in displacement during landslide damage. Our findings yield promising results. 

Fig.\ref{Figure6}b illustrates the impact of deformation measurements from other monitoring points on single target forecasting results for specific monitoring points in the \textbf{ST-NSP} scenario. The forecasting results reveal significant differences compared to the observed values, explaining the elevated prediction error metrics in the \textbf{ST-NSP} scenario and underscoring potential limitations in forecasting due to the unique landslide behaviors across different sections. Consequently, the interpretability for surrounding measurements offered by the LFIT model opens new avenues for understanding spatial relationships in landslide deformation forecasting. Extending this approach to forecasting scenarios involving multiple monitoring sites in the case of other natural hazards is possible, thus contributing to a more comprehensive and systematic understanding of these phenomena.

Furthermore, LFIT can estimate forecasting skills for all deformation measurements in multi-target forecasting scenarios incorporating location-specific information from monitoring point labels (\textbf{MT-MPC}). By analyzing the frequency distribution of various cumulative displacement states and examining the degrees of overlap between the blue and orange dots, as represented by the gray bars, we can better understand the areas where the interpretable neural networks demonstrate exceptional forecasting performance for multiple landslide sections. 

A comparison of Fig.\ref{Figure6}c shows that the model performs worse on deformation data that exhibit oscillatory patterns with fluctuations and abrupt changes. In contrast, the most accurate predictions are achieved for monitoring point data displaying a distinct step-like variation pattern. This observation is consistent with previous studies, which have demonstrated impressive prediction performance for step-like displacement series. In this work, without incorporating additional environmental variables or prior knowledge, we have identified a weakness in interpretable deformation forecasts, namely, the challenge of forecasting abrupt changes or displacements needing more obvious patterns.

\subsubsection{Interpretable Landslide Forecasting: Enhancing Neural Networks Temporal Awareness}
\label{Interpretable Landslide Forecasting: Enhancing Neural Networks Temporal Awareness}

The interpretable forecasting results of LFIT facilitate a comparison between observations and predictions, allowing for the assessment of their consistency and, thus, understanding the model's accuracy. Distinct from conventional forecasting results, the non-linear components derived from LFIT's interpretable attention blocks can be visualized. These weights embody the deep neural network's perception of temporal patterns and can represent the learned relevant patterns and temporal dynamics. This visualization allows for an evaluation of the physical consistency in the results obtained from the interpretable neural network.

\begin{figure*}[!ht]
	\centering
	\captionsetup{labelfont=bf}
	\includegraphics[width=16.5cm]{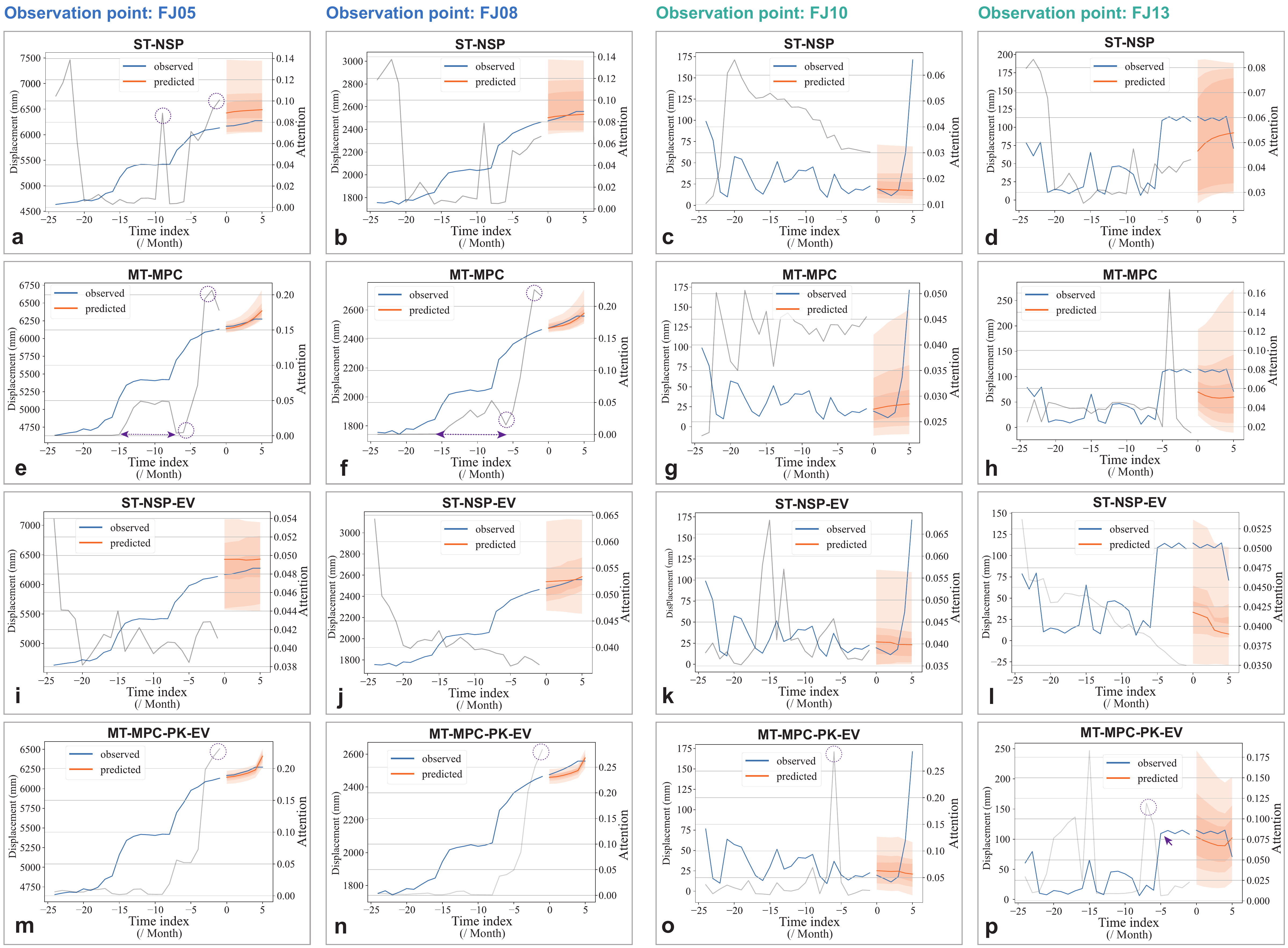}
	\caption{\textbf{Interpretable Forecasting results for Various Deformation Patterns at Observation Points across Different Scenarios.} \textbf{a-d.} Interpretable Forecasting results in \textbf{ST-NSP} scenario: single-target forecasting without considering external factors. \textbf{e-f.} Interpretable Forecasting results in \textbf{MT-MPC} scenario: multi-target forecasting using monitoring point prior categorical features without considering external factors. \textbf{i-j.} Interpretable Forecasting results in \textbf{ST-NSP-EV} scenario: single-target forecasting integrating external factors. \textbf{m-n.} Interpretable Forecasting results in \textbf{MT-MPC-PK-EV} scenario. Multi-Target Forecasting incorporating Prior Knowledge and external predictors. As spatially specific prior knowledge and environmental factors are integrated, deformation measurements from different monitoring sites exhibiting similar temporal patterns display corresponding extrapolated trends (orange line). For example, at two neighboring monitoring points with similar deformation trends (blue curves in m and n), highly similar forecast trends (orange curves) for the future six-time steps have been generated.} 
	\label{Figure7}
\end{figure*}

Fig.\ref{Figure7} illustrates the interpretable forecasting results for four experimental scenarios across different monitoring points (Fig.\ref{Figure7}a-d). In terms of forecasting results, after integrating prior knowledge, interpretable landslide forecasting demonstrates high accuracy and temporal awareness, even for deformation patterns exhibiting significant fluctuations (Fig.\ref{Figure7}e-h, \ref{Figure7}m-p). Furthermore, as more spatially specific prior knowledge and environmental factors are integrated, deformation measurements from different monitoring sites exhibiting similar temporal patterns display extrapolated trends with high consistency (orange line). This result indicates that the enriched context provides a global, physically sensible view of landslide behavior. Thus, other measurements could largely influence extrapolation in individual measurements, as these landslide monitoring components may share similar mechanisms. In this context, the fusion of prior knowledge and other predictor variables enables high-skilled landslide forecasting, even when the deformation lacks step-like regularity (e.g., Fig.\ref{Figure7}c, d).

Moreover, the predicted uncertainties can identify unreliable forecasts. In the case of a monitoring point with a sudden acceleration in landslide rates (Fig. \ref{Figure7}d), the increasing trend can be correctly forecasted. However, it is also associated with higher uncertainties, which suggests lower confidence in accurately forecasting deformation values under such rapid changes.

The interpretability of the above forecasting is expressed by the gray lines in the figure, which represent attention weights and denote the relative importance of each past input for predicting current landslide deformation. These weights reveal how the model prioritizes different periods of historical data when generating forecasts for the monitoring point under varied conditions, particularly in the \textbf{MT-MPC-PK-EV} forecasting scenario. The learned temporal patterns demonstrate meaningful spatial and temporal awareness. For instance, in measurements that exhibit a step-like deformation pattern, which has a relatively strong autocorrelation, the model relies on recent measurements (high attention weights for the preceding 3-5 days) when landslide deformation rapidly accelerates. This indicates that those time steps contain the most relevant information for predicting changes.

In contrast, during a period of slower deformation with minor fluctuations, the model incorporates data from a longer historical duration (greater attention for the preceding 5-10 days), demonstrating consideration of a broader context. Such attention-based mechanisms suggest that, when infusing spatial-specific knowledge into interpretable deep neural networks, they develop a nuanced, localized analysis of how historical landslide behavior at each monitoring point informs future forecasting across varied conditions. The neural network adapts physically sensibly based on how rapidly and dramatically conditions evolve at each monitoring point.

We also evaluate the extrapolative capabilities and interpretability of interpretable landslide forecasting using knowledge-infused LFIT for extended timescales. To achieve this, we increase the length of input sequences and expand the forecasting horizons. We investigate the model's ability to handle longer forecast lengths and incorporate additional input information, focusing on the \textbf{MT-MPC-PK-EV} scenario, where prior knowledge and influencing factor observations are included. We will gradually extend the forecast lead time to 12 months to assess long-term forecast skills. Meanwhile, we explore the influence of sequence length in historical landslide deformation data by progressively extending the historical data length to 48 months. We aim to determine the amount of temporal information the model requires to achieve accurate forecasts. We will consider two cases with input lengths of 24 and 48 months while evaluating two forecasting horizons of 8 and 12 months. In addition, we will add a priori time-varying information, such as year, month, and season, to the input to enhance the model's understanding of temporal dynamics, thereby capturing periodic variations inherent in Huanglianshu's landslide displacement sequences.

As anticipated, the landslide forecasting performance using knowledge-infused LFIT generally decreases with increasing forecast horizons. When employing 48 months of historical data and setting a 12-month forecast horizon, the error metrics between predictions and observations show substantial improvement across most monitoring points (Supplementary Figure). The effects of lengthening the input sequence vary by location. Longer inputs reduce forecasting accuracy for monitoring points in the reservoir level influence zone. This can be attributed to the fact that displacements in this zone predominantly depend on reservoir operations and are insensitive to longer-term trends. In contrast, longer inputs enhance predictions at rainfall-influenced sites, such as FJ-010 and FJ-013. This improvement may be due to more relevant information on historical environmental conditions and displacement responses. We assume that at rainfall-driven sites, the longer inputs enable the model to learn better complex relationships between precipitation, seasonal variations, and displacements, thereby boosting predictability. 

In the context of interpretability in the time dimension, we expect to understand which temporal pattern has been learned and if these contents from attention blocks of LFIT can have physical plausibility. We visualize and compare the temporal patterns extracted from input sequences of different lengths and focus on the \textbf{MT-MPC-PK-EV} forecasting scenario. The proposed interpretable Landslide Deformation Forecasting achieves a certain level of physical interpretability. 

\begin{figure*}[!ht]
	\centering
	\includegraphics[width=16.5cm]{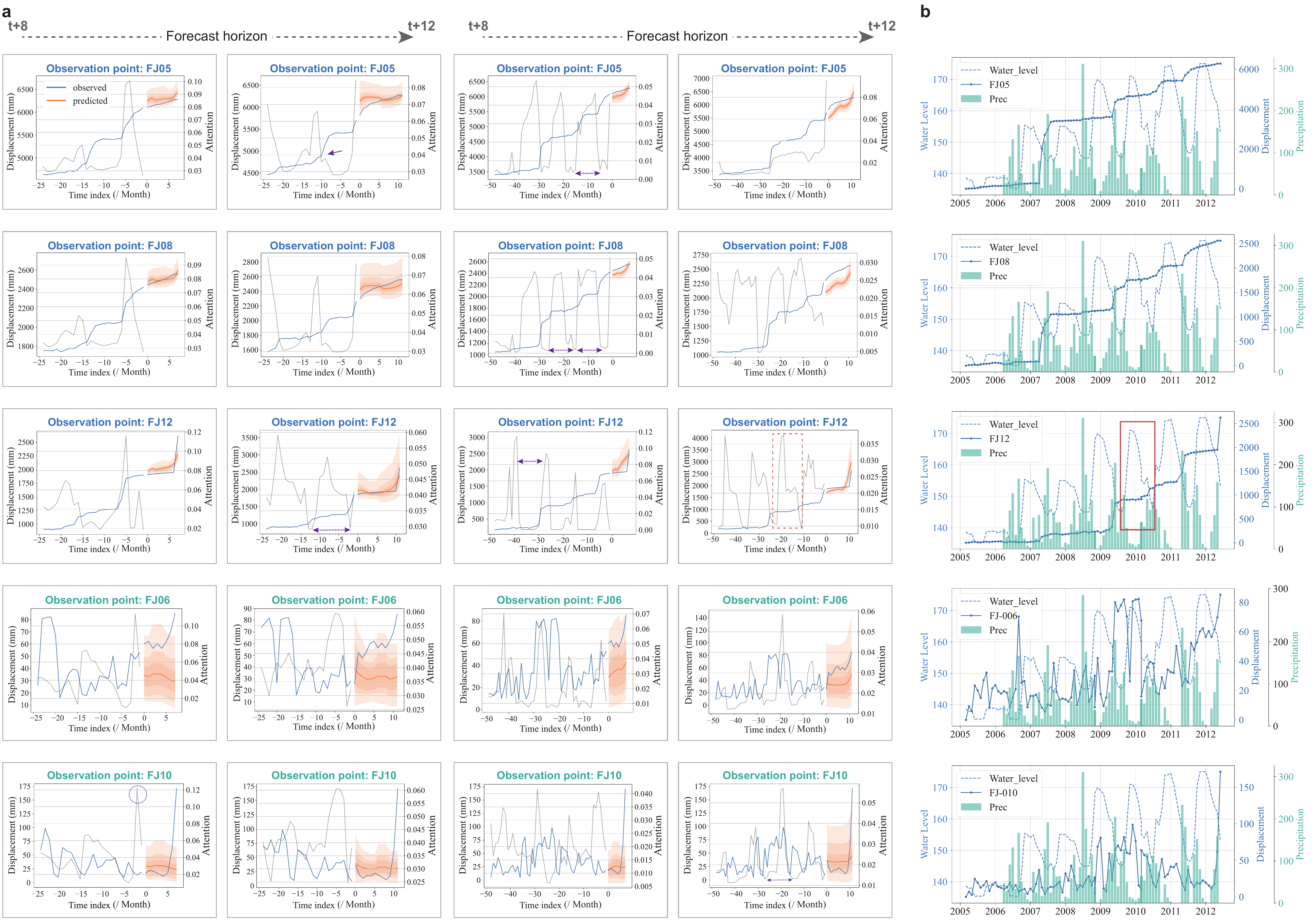}
	\caption{\textbf{Knowledge-Guided lFIT's interpretation of temporal correlations between water level changes and deformation in the \textbf{MT-MPC-PK-EV} forecasting Scenario.} \textbf{a.}  Interpretable Landslide Forecasting results for varying historical information and forecasting horizons across multiple observation points. \textbf{b.} Time series of landslide deformation, water level, and rainfall observations. In the \textbf{MT-MPC-PK-EV} scenario, LFIT effectively captures the relationship between periodic reservoir fluctuations and corresponding landslide displacement patterns, emphasizing rapid increases and cyclical trends. 48-month historical data appears to enhance the neural networks' recognization of intrinsic physical drivers.} 
	\label{Figure8}
\end{figure*}

The temporal correlation between periodic reservoir level changes and corresponding slope deformation is effectively captured by our model, as illustrated in Fig.\ref{Figure8}. The interpretable forecasting results reveal consistent features, including a notable emphasis on increased displacement (evident by the rising gray curve preceding deformation) and implicit period patterns. Notably, the learned cyclical pattern becomes more pronounced when 48 months of information are incorporated. Furthermore, the distribution of learned temporal characteristics exhibits remarkable regularity across different parts of the landslide body. At monitoring points near the leading edge of the landslide, the model accurately captures deformation fluctuations that closely align with the reservoir pattern, as demonstrated in Fig.\ref{Figure8}b. This alignment significantly enhances the accuracy of our forecasts. Notably, at monitoring point FJ12, our model displays an impressive correspondence between the learned periodic patterns and the actual water level fluctuation curve. This alignment ensures predictions that closely match the observed data and enables the early identification of rapid increases in deformation before slope failure. Despite gradually attenuating the periodic pattern with increasing distance from the leading edge, its presence can still be detected at most monitoring points. This finding is solid evidence that knowledge-guided LFIT methodology effectively captures the spatially heterogeneous response exhibited by this landslide.

Upon comprehensive comparison, we find that the learned periodic patterns exhibit characteristics of water-level changes, with approximately a 12-month interval between two peaks in a gray curve. This implies that the landslide leading edge region strongly correlates with water level variations within 12 months without lag time. A hysteresis effect between water level fluctuation and landslide displacement can explain this. Generally, the magnitude of the lag time directly corresponds to the permeability of the landslide body. It serves as an indicator of slope stability influenced by long-term reservoir water fluctuations. The cyclical fluctuations in the water level subject the landslide body to repeated cycles of water immersion and wind-induced drying. This reciprocating erosion of the seepage channels, driven by the periodic water level fluctuation, progressively damages the slope rock mass. During reservoir water level rises, water infiltration into the interior of the rock and soil body intensifies rock fracture interactions through water-chemical processes. As the reservoir level declines, new secondary pore spaces are generated within the slip body. Water outflow carries mineral particles away from fissures, creating new infiltration channels that facilitate subsequent water level rises. The long-term periodic fluctuation of the reservoir level impacts the infiltration channels within the landslide body, serving as conduits for water infiltration and discharge, thereby influencing slope stability. The cyclic rise and fall of the water level continuously scour and infiltrate the geotechnical body, gradually enlarging the permeability channels on the surface of the landslide body.

Consequently, water infiltration and discharge within the landslide body become more facilitated, gradually reducing the lag time between landslide displacement and reservoir water fluctuations, indicating a strong correlation where landslide deformation responds to water level fluctuation. Moreover, knowledge-guided LFIT effectively capture the phenomenon of step-like displacement during accelerated deformation, typically observed when the reservoir level decreases. Studies indicate that as the water level drops, internal water within the landslide flows outward, leading to seepage pressures pointing away from the slope, negatively impacting slope stability. Furthermore, the decrease in reservoir level diminishes the supporting force of water on the slope, accelerating the rate of slope deformation.

Overall, interpretable landslide deformation forecasting results demonstrate that Knowledge-infused LFIT effectively understands the lag correlation between reservoir level changes and displacement, reflecting reservoir-driven landslide theories. Utilizing at least 48 months of historical data as model input is reasonable. The significance of the learned periodic patterns becomes particularly evident when 48 months of data are incorporated as input. This observation implies that for complex, time-lagged systems such as reservoir landslides, utilizing longer input sequences enhances the ability of knowledge-infused LFIT models to capture the intrinsic physical drivers governing the system. In turn, these models can effectively unravel the fundamental processes that establish the connections between the predictor variable (reservoir level) and the predictand variable (landslide displacement). This valuable insight can transcend the limitations of models that rely exclusively on short-term correlations.

\subsubsection{Deciphering Influencing Factors: Unveiling the Drivers of Forecasting}
\label{Deciphering Influencing Factors: Unveiling the Drivers of Forecasting}

Given the distinctive capability of the LFIT to disentangle individual feature contributions and identify feature interdependencies, we can effectively isolate local contributions from the primary controls discovered by the LFIT. In the \textbf{ST-NSP-EV} scenario, regional analyses for assessing landslide drivers and controls reveal that the influence of factors varies across different parts of the landslide body. 

\begin{figure*}[!ht]
	\centering
	\includegraphics[width=16.5cm]{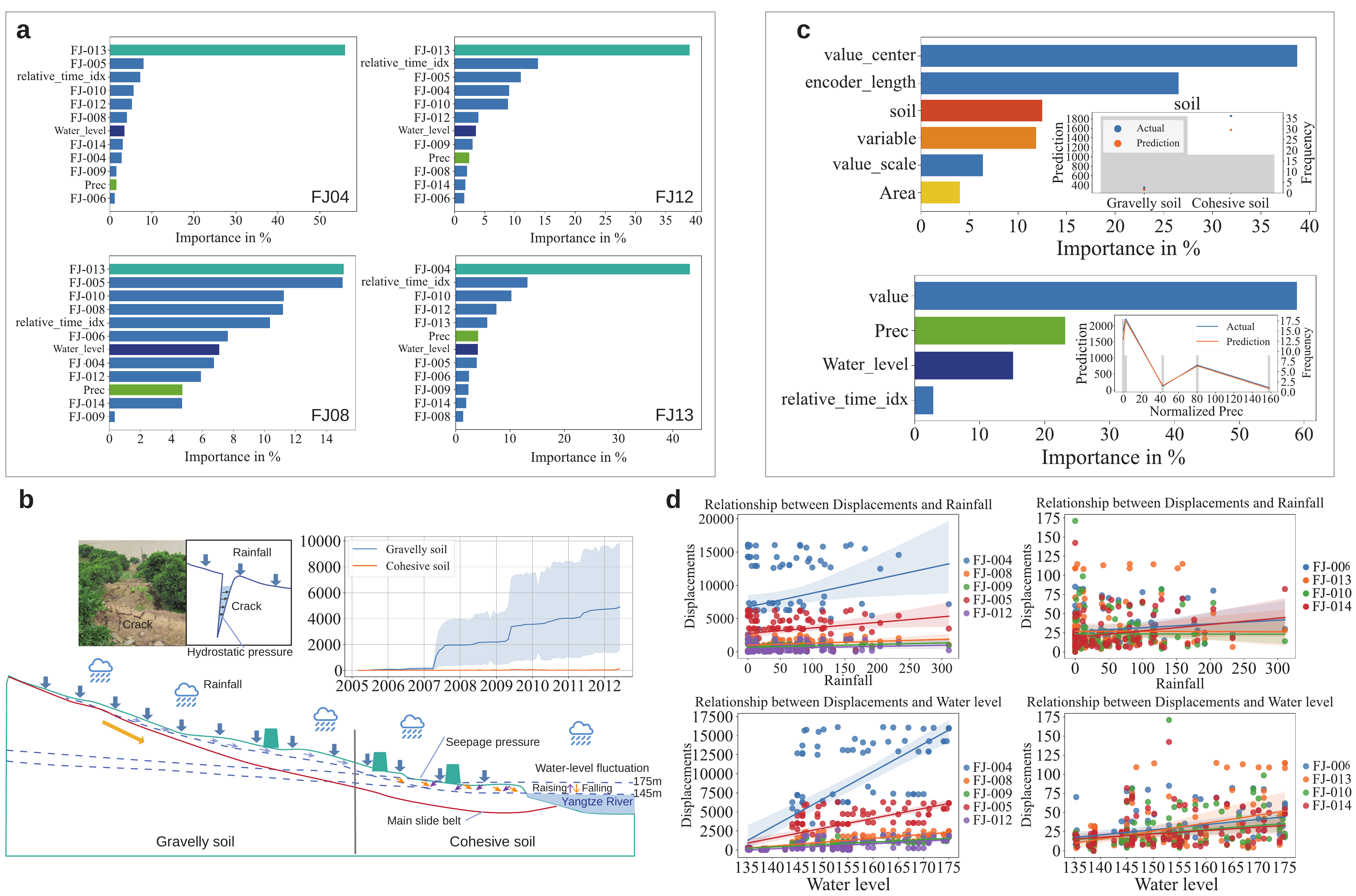}
	\caption{\textbf{Analysis of forecasting contributions from influencing factors the reservoir landslide. } \textbf{a.}  Dynamic observations (predictors) contribute to landslide deformation forecasting in the \textbf{ST-NSP-EV} scenario. \textbf{b.} Schematic Illustration of Rainfall and Water Level Effects on Different Areas of Water-Level Landslides. Reservoir Landslide Cross sections inspired by Figure 5 of \cite{055Yang2023}. Figure not to scale.\textbf{c.} Contribution of static attributes (a priori knowledge) and dynamic observations (predictors) to landslide deformation forecasting in the \textbf{MT-MPC-PK-EV} scenario.\textbf{d.} Scatter Plot of External Factors and Landslide Deformation.} 
	\label{Figure9}
\end{figure*}

As depicted in Fig. \ref{Figure9}a, at the leading edge of the landslide, the contribution of water level surpasses that of precipitation in terms of landslide deformation forecasting skill. Conversely, in the rear area of the landslide, precipitation exhibits a more significant contribution compared to the water level. These results align with the underlying physical processes. Fig.\ref{Figure9}b provides a visual representation of how variations in rainfall and water levels influence different areas within water-level landslide zones.  The leading edge of the landslide is near the Yangtze River region. The rising water levels immerse this region of the landslide body, leading to soil saturation and softening, which weaken the geotechnical body and can result in failure. Additionally, water infiltration increases the mass of the geotechnical body, further compromising its stability \cite{055Yang2023}. On the other hand, the central and trailing edges of the landslide are not in direct contact with the river water; therefore, their response to reservoir level changes is less pronounced. In these regions, rainfall infiltration exerts a downward force on the slip zone and reduces resistance, ultimately leading to deformation in the landslide parts \cite{060Ye2022}.

As shown in Fig.\ref{Figure9}c, in the \textbf{MT-MPC-PK-EV} scenario, we quantify the contribution of location-specific prior knowledge and several features that require standardization. Geotechnical properties play a more significant role than monitoring point labels. The attributes of hazardous areas, defined based on experience, do not contribute significantly to the prediction. When comparing contributions for time-dependent input, the deformation variables exhibit a more significant influence on the overall prediction, suggesting that the interaction between different deformation measurements substantially impacts the forecasting outcomes. Rainfall and water level follow in importance. Fig.\ref{Figure9}d demonstrates the correlation between external factors, rainfall, water level, and landslide deformation through a scatter plot based on observational data.

In Supplementary Figure, water level consistently emerges as the most crucial factor affecting prediction skills when setting various historical sequences and forecasting horizons. This finding elucidates why the learned periodic patterns closely resemble water level fluctuations. Even when historical sequences and forecasting horizons differ, geotechnical properties consistently exert the most significant impact on landslide forecasting results. 

Based on the above analysis, LFIT enables identifying crucial factors and quantifying their potential contributions to landslide predictability, emphasizing the significant role of strong slope-climate coupling in landslide occurrences. The composite features, such as water level and precipitation, identified by the Knowledge-infused LFIT align with existing understandings of reservoir landslide mechanisms. These connections were not explicitly demonstrated in previous data-driven landslide deformation forecasting studies.

\subsection{Forecasting deformation and Deciphering Influencing Factors for Landslide with deep-seated creep deformations}
\label{subsec:3:2Forecasting deformation and Deciphering Influencing Factors for Landslide with deep-seated creep deformations}

Our second study case is in the Tibetan Plateau, a region renowned for distinctive geological and meteorological features. However, the equilibrium of this extensive plateau is being disrupted due to the intensification of human activity and the far-reaching consequences of climate change. As a result, the increasing prominence of landslide risk poses a growing threat. Landslides in this area typically develop from highly fractured slopes, exhibiting ongoing slow movements, as evidenced by deposits along valleys. These fragile slopes are susceptible to external triggers such as earthquakes and rainfall.

The 2018 Baige landslide, a representative example of landslides characterized by long-term deep-seated gravitational slope deformations culminating in a sudden collapse, occurred within a high-stress suture zone. This zone was subject to rapid river erosion and weathering-induced instability combined. Originating subtly in 1966, this landslide displayed an extensive history of slow yet significant deformation. Like other landslides exhibiting long-term creeping deformation, the Baige landslide culminated in catastrophic outcomes upon sudden acceleration, blocking the Jinsha River twice. These events underscore the urgency to understand and predict the evolution of such hazards. Enhancing the predictability of these large, slow-moving landslides is vital for risk mitigation to human life and infrastructure.

\begin{figure*}[!ht]
	\centering
	\captionsetup{labelfont=bf}
	\includegraphics[width=16.5cm]{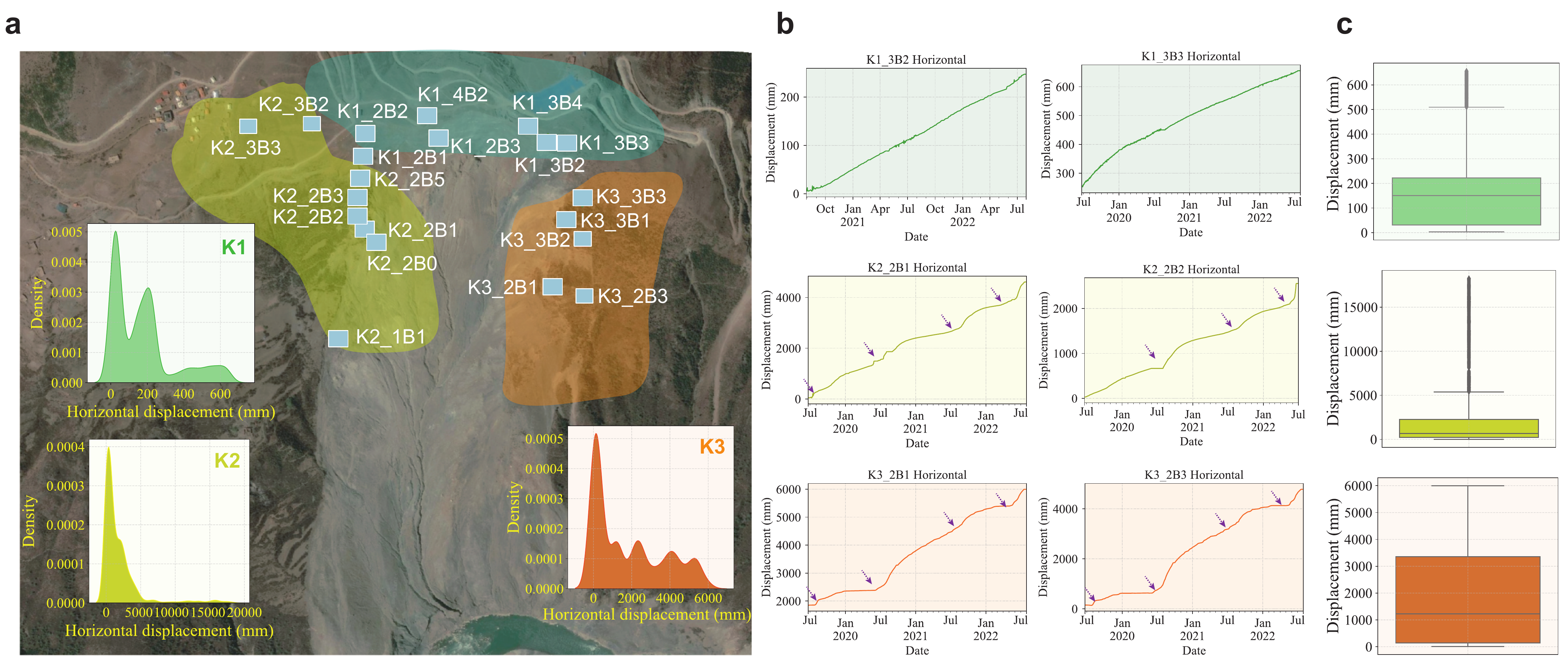}
	\caption{\textbf{Slow-moving Landslide Cases.} \textbf{a.}  The 2018 Baige landslide in the Tibetan Plateau's high-stress suture zone. \textbf{b.}  Seasonal deformation patterns in the three unstable zones (K1, K2, and K3) of the Baige landslide. \textbf{c.} Deformation variables.} 
	\label{Figure10}
\end{figure*}

The Baige landslide experienced continuous deformation following two significant collapses in 2018, with accelerated movement observed during the summer months (Fig.\ref{Figure10}). As shown in Fig. 8b, the deformation pattern diverges within three unstable zones, namely K1, K2, and K3. The K1 zone demonstrates consistent displacement over time, while the K2 and K3 zones experience abrupt surges each summer, particularly in June before the onset of the rainy season. Both annual and seasonal fluctuations in landslide deformation are evident (Supplementary Figure), with cumulative movement increasing after the 2018 collapse. The K1 zone exhibits a linear increase in deformation, while the K2 and K3 zones remain relatively stable from October to May but then experience rapid acceleration after that. The K3 zone exhibits the most significant deformation. The timing of accelerated movement corresponds with the arrival of monsoonal rains, suggesting that changes in pore-water pressures triggered by precipitation are likely responsible for these observations \cite{066Handwerger2022}.

Here, we aim to enhance the predictability of such large-scale landslides on the Tibetan Plateau by incorporating prior knowledge into an interpretable neural network. This enables interpretable forecasting of landslides in this important landslide-prone area. By leveraging the region's unique geological characteristics and prevalent adverse factors, our investigation seeks to elucidate the complex interplay among these factors and the potential for sudden landslide deformation accelerations. We explore three experimental scenarios to examine the influence of prior knowledge and environmental variables on large landslide forecasting.

In the single-target forecasting with environmental variables (\textbf{ST-NSP-EV}) scenario, we incorporate time-dependent covariates representing influential factors, such as elevation, slope, humidity, temperature, and rainfall, to predict the displacement of an individual landslide. The model captures the temporal variability in landslide behavior by including these dynamic variables, adding coordinate information as additional time-dependent covariates.

In the multi-target forecasting with monitoring points as categorical information (\textbf{MT-MPC}) scenario, we utilize multidirectional displacement time series as categorical data and monitoring points as an additional class label to represent localized landslide conditions. This approach offers a spatiotemporal understanding of diverse landslide behaviors across space.

The multi-target forecasting with prior knowledge and environmental variables integrated with monitoring point categorical information analysis (\textbf{MT-MPC-PK-EV}) scenario builds upon the \textbf{MT-MPC} scenario by incorporating additional prior information, including both static properties of the displacement data, such as characteristics of unstable regions and time-dependent influential factors like elevation, temperature, humidity, and rainfall.

\begin{figure*}[!ht]
	\centering
	\includegraphics[width=16.5cm]{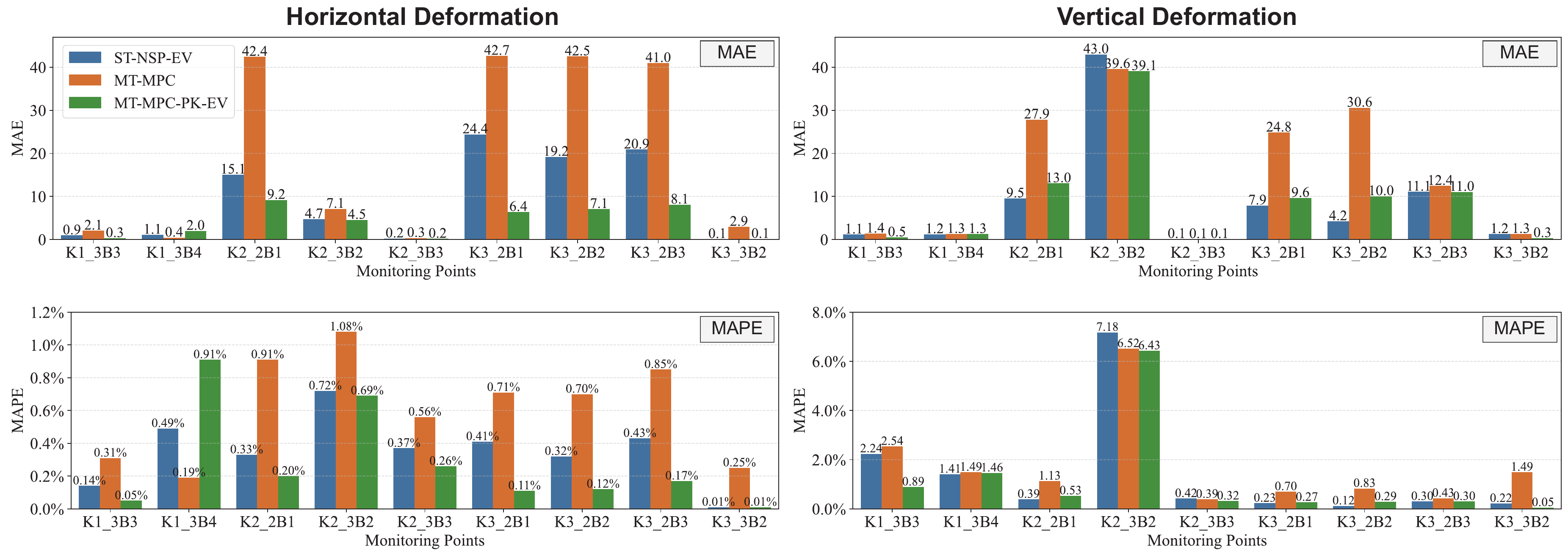}
	\caption{\textbf{Deformation Forecasting Error Metrics in Different Scenarios across Unstable Zones.}} 
	\label{Figure11}
\end{figure*}

First, we examined the error metrics of deformation forecasting in different scenarios across three unstable zones. As shown in Fig.\ref{Figure11}, Consistent with our expectations, integrating multiple environmental predictors (\textbf{MT-MPC-PK-EV} scenarios) and a priori knowledge significantly improved forecast precision. For unstable zones such as K2 and K3, the forecasts skill of LFIT is substantially enhanced when incorporating additional prior knowledge and environmental observations. Conversely, including this information adversely affected the forecast results for deformation in the K1\_3B4 monitoring points. This observation suggests that the forecasting skill of LFIT is heavily influenced by heterogeneous landslide regions, which is understandable given their complex behavior, governed by a multitude of diverse factors.

\begin{figure*}[!ht]
	\centering
	\captionsetup{labelfont=bf}
	\includegraphics[width=16.5cm]{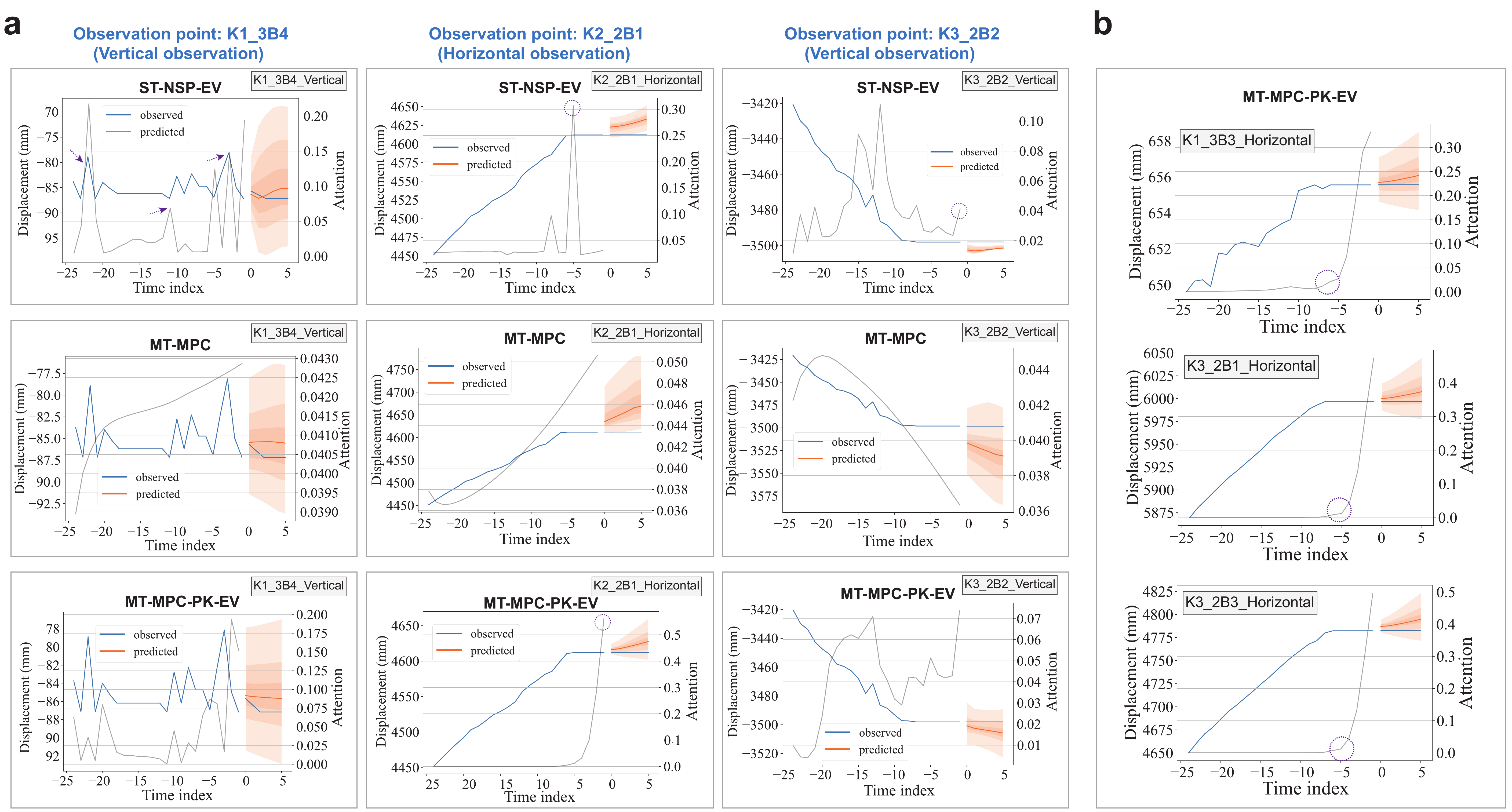}
	\caption{\textbf{Interpretable Forecasting results for Various Deformation Patterns at Observation Points across Unstable Zones.} \textbf{a.}  Improved predictability of LFIT for landslide deformation in horizontal and vertical directions by integrating multiple environmental predictors and spatial-related prior knowledge. The grey line represents the neural network's temporal awareness, focusing on the present time step during forecasting and suggesting higher confidence in predicting current trends.  \textbf{b.} Learned temporal pattern of the knowledge-infused LFIT, showing a significant rise at a lag of five-time steps in monitoring sites K1 and K3 (right landslide). This pattern indicates that the landslide may remain unchanged over the forecasted six days, with larger displacements possible over longer timescales.} 
	\label{Figure12}
\end{figure*}

Then, we investigated the forecasting results of LFIT for landslide deformation in horizontal and vertical directions. As hypothesized, integrating multiple environmental predictors (\textbf{MT-MPC-PK-EV} scenarios) and spatial-related priori knowledge improved the predictability for this complex landslide. According to the grey line, which represents the neural network's temporal awareness and landslide behavior's temporal features, we find that the knowledge-infused LFIT focuses strongly on the present time step during forecasting, suggesting higher confidence in predicting current trends (Fig. \ref{Figure12}). Specifically, the learned temporal pattern exhibits a significant rise at a lag of five-time steps in monitoring sites K1 and K3 (right landslide) (Fig. \ref{Figure12}b). This pattern indicates that the landslide may remain unchanged over the forecasted six days, although larger displacements are possible over longer timescales. Reports of the sudden collapse of the Baige landslide in December 2022 support this speculation. We thus think that the temporal awareness from knowledge-infused LFIT is physically justified. 

Supplementary Figure presents LFIT's ability to identify periodic patterns in 3-hourly observations of Baig landslide deformation, demonstrating LFIT's temporal awareness and capacity to comprehend the inherent temporal dynamics associated with large landslide deformation.  

Therefore, although achieving high-precision forecasts remains challenging due to the complex nature of the Baige landslide and the omission of external factors (e.g., earthquakes), our results represent meaningful progress. With added a priori knowledge, the neural network model demonstrates promising physical consistency in temporal perception, which could guide future modeling of landslide behavior.

Further, the variable selection results of LFIT allow us to discern the impact of various predictors, including environmental variables and surrounding deformation observations. Large landslide forecasting allows for an assessment of the comprehensive influence of internal and external factors on localized landslide behavior. 

\begin{figure*}[!ht]
	\centering
	\captionsetup{labelfont=bf}
	\includegraphics[width=16.5cm]{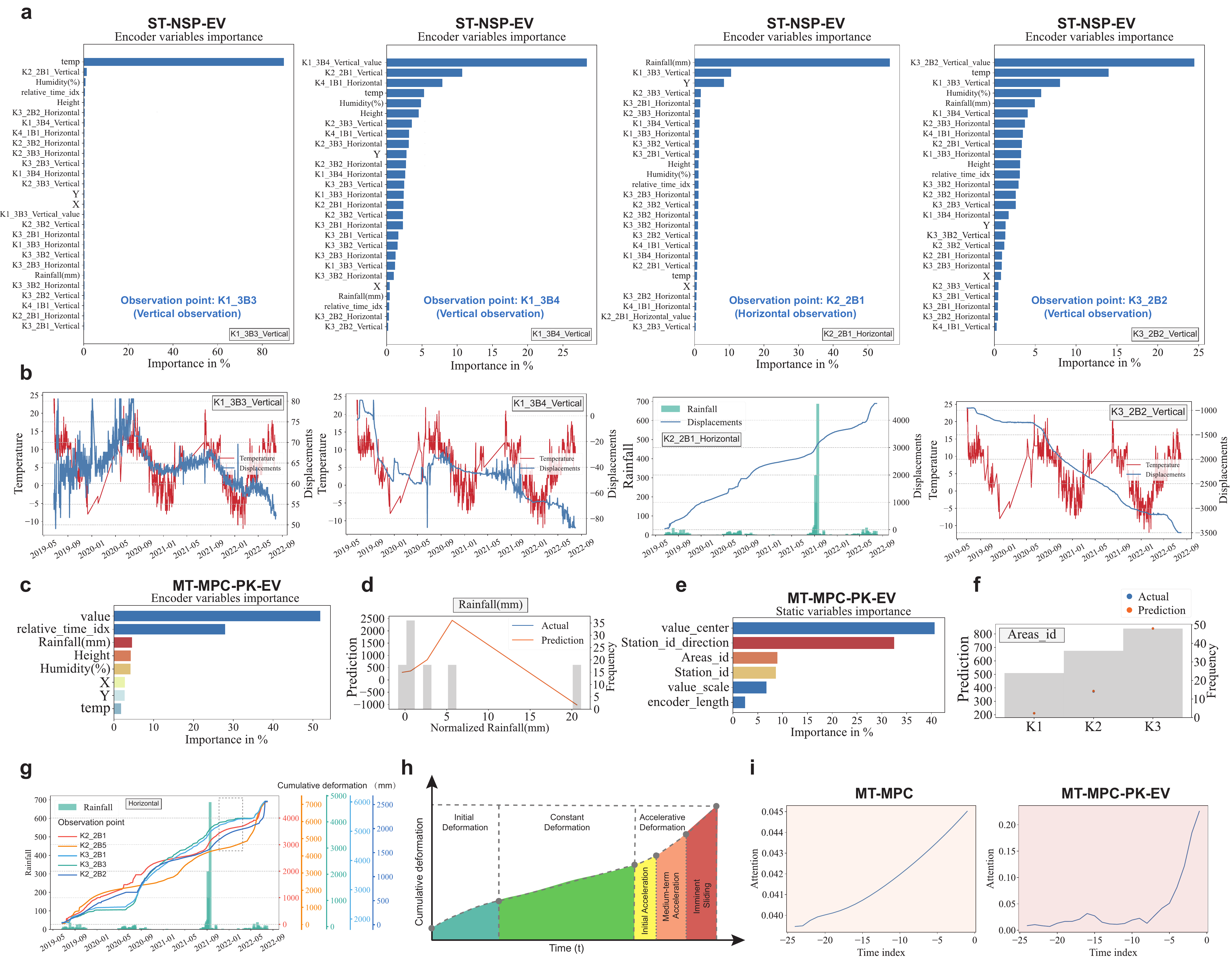}
	\caption{\textbf{Analysis of Forecasting Contributions from Influencing Factors in the large-scale landslide} \textbf{a.} Variable Selection Results and Contributions to LFIT Forecasting in the \textbf{ST-NSP-EV} Scenario.   \textbf{b. }Correlations between environmental variables and deformation across monitoring sites. \textbf{ c-e. } Variable Selection Results and Contributions to LFIT Forecasting in \textbf{MT-MPC-PK-EV} Scenarios. \textbf{g.} Relationship between deformation patterns and rainfall in the K2 and K3 zones.\textbf{ h. }Standard creep curve of landslides (revised after Saito, [1969]). \textbf{i.} Temporal Features Learned by LFIT's Interpretable Attention Blocks.   }
	\label{Figure13}
\end{figure*}

In the \textbf{ST-NSP-EV} scenario, our analysis reveals the significant contribution of environmental variables and other deformation observations to the local forecasting of landslide behavior, thus identifying the specific factors that influence the behavior of landslides in particular areas (Fig. \ref{Figure13}a). The results reveal that the contributions of predictor variables exhibit significant variation across different landslide sections, further elucidating the distinct mechanisms at play within various landslide parts. For instance, when examining the effects of external factors, we observed that LFIT relied predominantly on rainfall in certain landslide sections. Conversely, temperature or soil moisture played a more prominent role in others, findings consistent with our data analysis. The close association between rainfall and deformation was prominently evident across most monitoring sites in the K2 and K3 zones. Specifically, our findings reveal that LFIT forecasts in the K1 region were particularly sensitive to temperature, as demonstrated by the strong correlation between temperature and monitoring points within the K1 region (Fig. \ref{Figure13}b). This relationship was effectively captured by LFIT, leading to improved predictions. Likewise, in the K2 zone, the model identified rainfall as the primary driver, exerting the most significant influence on the prediction.

In the \textbf{MT-MPC-PK-EV} scenario, LFIT further enhanced predictability by emphasizing rainfall as the predominant factor governing the overall prediction (Fig. \ref{Figure13}c). By exploring interpretability in spatial-related prior knowledge, LFIT's interpretability revealed potential limitations in predictive accuracy, specifically within the K2 zone. The K2 zone may experience rare events, such as heavy rainfall, which could reduce the predictability of landslide behavior in this area. Thus, this limitation may arise from LFIT's difficulty in capturing infrequent occurrences without incorporating additional information. 

Above all, according to feedback from the interpretable neural network LFIT, in Baige landslide forecasting, precipitation and temperature emerge as dominant controls, although their relative importance varies across different landslide parts. Recent studies suggest that warming will likely increase the probability of landslides in the Tibetan Plateau area. As the impacts of climate change continue to intensify, incorporating knowledge into large landslide forecasting in this region using interpretable neural networks will further improve landslide predictability and advance our understanding of the intricate interactions between landslides and environmental factors, such as temperature, at regional and local scales. 

On the other hand, as illustrated in Fig. \ref{Figure13}i, the temporal features learned by the interpretable attention blocks of LFIT demonstrate an increasingly rapid growth trend in both the \textbf{MT-MPC} and \textbf{MT-MPC-PK-EV} experimental scenarios, where input encompasses spatial prior knowledge. This accelerated characteristic aligns with the third stage in the standard creep curve. The emergence of accelerated deformation signals during the tertiary creep stage holds significant implications for the effective functioning of landslide early warning systems. The subsequent collapse of the Baige landslide in December 2022 further validates that the temporal characteristics learned by the knowledge-infused LFIT have the potential to presage impending hazards. This underscores the immense potential of interpretable forecasting, architected by knowledge-infused neural networks, in facilitating early warning systems for landslides.

\section{Discussion }
\label{sec:4: Discussion }

Recent studies have highlighted the significance of incorporating domain knowledge in landslide modeling and forecasting. The integration of geological, geotechnical, meteorological, and terrain-related information fosters a comprehensive understanding of the factors driving landslide behavior. Our research builds on this growing body of knowledge by utilizing a transformer-based deep neural network architecture to develop a Knowledge-infused Interpretable Landslide Deformation Forecasting pipeline, demonstrating that incorporating prior knowledge into interpretable neural networks can enhance the predictability and interpretability of landslide behavior, and allowing a deeper exploration of the underlying landslide mechanisms based on the non-linear insights derived from the neural network's perception.

First, by infusing landslide-specific, regionally-related prior knowledge into the transformer-based architecture, the proposed forecasting pipeline enables the learning of complex, high-level non-linear relationships among regional conditions in specific landslide observational sites, corresponding landslide dynamics measurements, and external drivers. Such relationships represent, to some extent, the intricate interdependencies between diverse internal and external factors contributing to landslide evolution. The transformer-based architecture captures these relationships at various spatial and temporal scales. Regarding temporal aspects, local and global learning mechanisms capture sudden changes and periodic patterns in landslide behavior. Spatially, global spatial correlation learning aids in understanding the overall system, further capturing acceleration states for individual measurements without any precursors. Local learning captures landslide behavior in response to influencing factors at specific sites. The proposed knowledge-infused deep learning approach enhances landslide predictability by learning the underlying processes governing systematic landslide evolution under various influencing factors from a more comprehensive perspective.

Second, by explicitly exhibiting non-linear content, such as weights representing neural network perceptions from the transformer-based architecture, the proposed forecasting pipeline can elucidate dynamics features latent in time-dependent measurements that document landslide behavior, also decipher intricate interdependencies between diverse internal and external factors contributing to landslide occurrence. In terms of temporal aspects, the temporal patterns that recognized by interpretable neural network here unveil landslides' underlying dynamics under external influencing factors and identify future evolution trends. The ability to recognize temporal patterns and dynamics allows for the implementation of proactive measures to mitigate the potential impact of future events. Furthermore, our approach enables a deeper understanding of factors triggering landslide behavior by identifying significant influencing predictors. By illustrating temporal features in landslide behavior and pinpointing essential factors contributing to local and global landslide behavior, the proposed interpretable landslide forecasting pipeline can enhance interpretability of landslide forecasting and provide science understanding of landslide evolution.

\subsection{Infusing Knowledge into data-driven landslide forecasting has significantly improved our understanding of the complex dynamics governing landslide behavior}
\label{subsec:4:1 Infusing Knowledge into data-driven landslide forecasting has significantly improved our understanding of the complex dynamics governing landslide behavior}

In this study, we underscore the importance of forecasting landslide behavior at a systematic level rather than concentrating solely on site-specific observations. On the base of conducting a comprehensive analysis of multiple measurements across entire monitoring regions, we propose incorporating relevant spatial-related prior knowledge, thus achieving significant improvements in landslide predictability from a holistic perspective. 

Our proposed framework, Knowledge-Infused Landslide Forecasting using Neural Networks, comprises three key elements: (1) We examine landslide deformation across the entire monitoring region by simultaneously taking measurements from multiple sites. This approach enables us to understand and forecast landslide behavior more comprehensively. (2) We incorporate spatially-related information from documents and expert insights as prior knowledge to represent local conditions at specific observational sites of landslide deformation. This infusion of prior knowledge enhances the spatial awareness of data-driven landslide forecasting methods, facilitates entire landslide forecasting, and advances a holistic understanding of landslide evolution. (3) We employ a transformer-based deep network architecture to integrate data and knowledge effectively. This neural network was renowned for its multimodal information fusion and non-linear feature learning capabilities. It can capture the intricate dynamics of landslides influenced by internal and external factors through unparalleled non-linear mapping techniques.

The three key points are based on the following principles and assumptions.

First, we conceptualize landslides as interconnected networks of discrete patches, where each patch influences the others \cite{065Lim20211748}. According to our hypothesis, observations of landslide behavior at specific monitoring sites can serve as proxy indicators of specific patches within the landslide system \cite{066Handwerger2022}. The relationships between these observations represent the interconnections between patches, enabling the forecasting model to identify the overall behavior of the landslide system and even detect abrupt changes in landslide behavior in relatively stable areas. For instance, when landslides manifest precursor signals, such as a significant increase in deformation measurements at multiple monitoring sites before the catastrophic collapse, the deep neural network here effectively perceives these precursors by considering the interactions among patches. This enables the model to account for these signals when forecasting deformation at monitoring sites that lack apparent indicators of impending landslides. With this information transmitted and enhanced spatial awareness, our pipeline's deep neural network LFIT can effectively detect unforeseen situations at some monitoring sites that would otherwise be difficult to identify and forecast based merely on observations from individual monitoring points. Furthermore, we incorporate observations of landslide drivers from additional data sources, which facilitates the differentiation of responses of different patches to various influencing factors. This ultimately makes large and complex landslides more predictable and provides a more comprehensive understanding of how they behave holistically. 

Second, to enhance the neural network's comprehension of the complex interactions between patches within the landslide system, we incorporate static attributes that provide the geographical context of each monitoring site where landslide behavior is observed, as prior knowledge. Typically, these location-related prior knowledge are qualitative descriptions in text. To incorporate this information into the neural network, we employ word embeddings, a text encoding method widely used in deep learning. Word embeddings facilitate the mapping of textual information into a high-dimensional space, leading to valuable insights into the underlying processes. Prior research has demonstrated the effectiveness of word embeddings in capturing intricate relationships between geological attributes \cite{067Padarian2019177,068Lawley2022}. Inspired by these findings, we leverage qualitative descriptions of the surroundings at each monitoring site as a form of prior knowledge, utilizing word embeddings to encode this descriptive information. By mapping these attributes onto a high-dimensional space of d dimensions, we successfully integrate prior knowledge into LFIT, enhancing its spatial awareness. Such a knowledge-guided deep learning model may consider local conditions and circumstances at each monitoring site, further improving the predictive capability for landslide behavior. 

Third, we employ a transformer-based deep neural network architecture with distinctive characteristics that enhance learning for the relationship between encoded prior knowledge and observed dynamics. This approach addresses the complexity of landslides by effectively capturing salient information while disregarding irrelevant details. Unlike shallow models, our LFIT employs a hierarchical structure, specifically utilizing GLUs to systematically identify higher-order interactions crucial for determining landslide dynamics. Gating in LFIT enables identifying local data and factors controlling behavior, exploiting spatial heterogeneity and attribute relationships often overlooked by shallow models. Additionally, incorporating LSTM strengthens our understanding of temporal dynamics. Previous studies have applied LSTMs directly to landslide displacement time series for prediction; however, our approach uses LSTMs guided by a priori knowledge from other neural network blocks, thus benefiting from insights gained from dedicated network components. By conditioning LSTMs with prior knowledge representation and initializing the LSTM cells and hidden states, LFIT captures the effects of intricate factor interactions on landslide kinetic behavior.

In summary, we leverage a fusion of knowledge and data to comprehensively represent deformation behaviors related to landslides, external trigger influences (e.g., water level), and static property information. This rich contextual understanding allows subsequent neural network components to uncover complex relationships within a deep, multifunctional architecture.  By capturing complex interactions among multiple inputs, Our proposed knowledge-guided landslide forecasting pipeline surpasses traditional approaches that focus solely on individual landslide behavior-related measurements. By integrating all relevant information, we can identify critical influences on landslide behavior at specific sites and at different stages, as well as discern how various landslide parts respond to external factors. Consequently, the ability to capture both heterogeneity and systematic patterns in landslide occurrences could be enhanced.
Nevertheless, landslide predictability will remain a significant challenge due to the inherent complexity of landslides and the potential influence of external factors such as earthquakes. Future research aims to incorporate additional types of a priori knowledge and expand the range of environmental variables considered influencing factors. For instance, integrating data sources such as microseismic signals and fracture monitoring could provide valuable insights into landslide dynamics when combined with existing knowledge embeddings. This would facilitate a more comprehensive understanding of the complex interactions that drive landslides. More approaches involve incorporating real-time data streams, such as rainfall patterns, groundwater levels, or satellite imagery. Recent studies in related fields, such as natural disaster forecasting, have demonstrated the benefits of incorporating a wide range of data sources and knowledge.

\subsection{Interpretable deep learning for Insights into Landslide Behavior: Trends and influencing factors}
\label{subsec:4:2 Interpretable deep learning for Insights into Landslide Behavior: Trends and influencing factors}

In this study, we present an interpretable landslide forecasting pipeline, which leverages a knowledge-infused deep neural network to reveal learned non-linear content. This distinguishing feature distinguishes our method from conventional data-driven models and post-event explainable approaches. Our interpretable landslide forecasting approach offers two key advantages. Firstly, in a self-explanatory manner, deep neural networks can directly illustrate their priorities and the critical information they rely on to produce extrapolation results, making the process transparent. Second, by examining the content recognized by interpretable deep learning, we can better understand the intricate interplay between external influencing factors and landslide occurrences at local scales. 

Technically, the attention mechanism, combined with a softmax activation function for weighing variables, plays a pivotal role in our methodology. It enables our model to capture and learn the complex patterns of landslide changes under the influence of multiple factors. Through the analysis of output weights and attention scores, we achieve interpretability, uncover latent temporal features hidden in the measurements of landslide behavior, and identify those factors critical for accurate forecasting, resulting in an increased understanding of the complex dynamics governing the behavior of landslides.  

Attention blocks can generate scores for each landslide behavior measurement value at every past time step. This capability enables us to identify past landslide behavior measurements with varying influence, highlighting their significance in accurate forecasts. Consequently, salient patterns in historical landslide behavior observation can be found, which can represent temporal features such as cycles and anomalies. By incorporating prior knowledge related to the condition of monitoring points in the surrounding areas, how the predictor variable influences the predictand variable through effects based on different landslide properties in various parts can be revealed. This capability enhances a better understanding of landslide dynamics and uncovers how landslide dynamics respond to external factors by utilizing learned temporal patterns as evidence. Variable selection blocks with a softmax activation can capture the intricate web of dependencies among various input variables, thus uncovering the effective combinations of factors that predominantly impact landslide displacement changes monthly or yearly.
Our comprehensive experimental analysis, encompassing various types of landslides, reveals the significant advantages of interpretable landslide forecasting.

In the investigation of reservoir landslides, the interpretable neural network uncovers a strong connection between the observed periodic deformation patterns on the landslide's leading edge and fluctuations in water level. This finding emphasizes the preeminent role of water level as the principal factor influencing landslide forecasting. We believe that our interpretable and knowledge-guided landslide forecasting approach discerns the impact of water level variations on slope deformation by capitalizing on the combined power of data and a priori knowledge. By comparing predictor variable (water level) measurements and learned temporal patterns for the predictand variable (deformation), we demonstrate that infusing prior knowledge into interpretable deep learning can capture the essential processes associated with reservoir landslides and that physically interpretable results can be attained. 

In the investigation of slow-moving, large-scale landslides, the interpretable neural network uncovers the crucial roles of rainfall and temperature in driving rapid deformation across distinct regions of the landslide. However, a holistic forecasting approach incorporating additional prior knowledge reveals that rainfall emerges as the most influential factor. This observation offers insights into the understanding of the Baige landslide in the Tibetan Plateau, an area subject to ongoing debates regarding its formation and causative factors. InSAR observations preceding the landslide event present persuasive evidence corroborating the idea that the slope experiences accelerated deformation during the rainy season [63]. Additionally, the significance of unstable area classifications is recognized, given that different regions display unique fissure distributions and rock properties. We postulate that knowledge-guided neural network LFIT appears to learn lithologically controlled weathering patterns, leading to rapidly increasing deformation trends as reflected in the attention weight results. Notably, this hypothesis was corroborated when the landslide collapsed three months after computing these findings. The Tibetan Plateau faces the widespread effects of climate change, leading to perceptible alterations in climate patterns and hydrological processes. Identifying the correlation between intrinsic slope properties and the evolving climate is essential for accurately predicting landslide behavior amidst these shifting conditions. By incorporating prior knowledge into deep learning models, our research methodology enables interpretable predictions of landslides, providing insights into the development and associated risks of landslides in these areas.

Future research endeavors will examine the applicability of interpretable predictions across a broader spectrum of landslide types and proxies that delineate their behavioral parameters. Recent research has emphasized the importance of coupling machine learning models with process-based models to better represent complex interactions between environmental factors and landslide occurrences. Integrating this approach in future studies could enhance predictive capabilities, incorporating data-driven insights and a physically-based understanding of landslide phenomena.

\section{Conclusion}
\label{sec:5: Conclusion}

In conclusion, our proposed interpretable landslide forecasting pipeline, which employs a transformer-based network called LFIT, can more holistically forecast landslide behavior. By incorporating landslide-specific, regionally-related prior knowledge into the transformer-based architecture, our pipeline can learn complex, high-level nonlinear relationships among regional conditions, landslide dynamics measurements, and external drivers, thus capturing the intricate interdependencies between diverse factors contributing to landslide evolution at various spatial and temporal scales. Therefore, interpretable landslide forecasting based upon knowledge-infused processes is capable of (1) revealing dynamics latent in time-dependent measurements that document landslide behavior, illustrating temporal characteristics of landslides, and (2) deciphering intricate interdependencies between diverse internal and external factors contributing to landslide occurrence, thus identifying essential factors that contribute to local and global landslides. In the study of reservoir landslides, we found that reservoir level fluctuations were the most significant influence on the location of the observation point at the leading edge of the landslide. At this point, the hidden periodic time-varying pattern coincides closely with these fluctuations, reflecting the underlying physical process behind reservoir-induced landslides. In the case study of a creeping landslide, we observed a rapidly increasing trend in deformation patterns identified from the measurements of creeping displacements, suggesting a potential for an imminent, significant shift in landslide behavior. This inference was later confirmed by the area's collapse just four months after our observation. The rainfall effect for landslide localized instability may cause such behavior. Our future work intends to enrich our interpretable landslide forecasting pipeline by integrating a broader range of knowledge and measurements about various landslides.

\section*{Suplementary}
\subsection*{Suplementary Text: Matrices to Measure Forecasting Performance}
\label{Suplementary Text 1: Matrices to Measure Forecasting Performance}
Matrices used in our study to measure the performance are:

Mean Absolute Error (MAE) represents the average of the absolute differences between predicted and actual values, measuring the error between model predictions and actual results. 

$$
\mathrm{MAE}=\frac{1}{n} \sum_{i=1}^n\left|y_i-\tilde{y}_i\right|
$$

Mean Absolute Percentage Error (MAPE) measures the average percentage difference between predicted and actual values, reflecting the relative degree of prediction error. However, zero values in the data can impact MAPE calculations.

$$
\mathrm{MAPE}=\frac{1}{n} \sum_{i=1}^n \frac{\left|y_i-\tilde{y}_i\right|}{y_i}
$$

Root Mean Square Error (RMSE) is the square root of the average squared differences between predicted and actual values. It is commonly used to assess a model's prediction error variance. Due to its sensitivity to outliers, using RMSE may result in inflated error values when data contains outliers. 

$$
\mathrm{RMSE}=\sqrt{\frac{1}{n} \sum_{i=1}^n\left(y_i-\tilde{y}_i\right)}
$$

Symmetric Mean Absolute Percentage Error (sMAPE) is a symmetric percentage error metric that reflects the deviation between predicted and actual values. It accounts for cases where predicted values are lower or higher than actual values, thus avoiding biased errors due to directional differences.

$$
\mathrm{sMAPE}=\frac{100 \%}{n} \sum_{t=1}^n \frac{\left|y_i-\tilde{y}_i\right|}{\left|y_i+\tilde{y}_i\right| / 2}
$$

Higher values for these four metrics indicate higher errors, suggesting unsatisfactory predictions, while lower values denote reduced errors and improved predictions.

\FloatBarrier 
\newpage
\section*{Acknowledgments}
This research was jointly supported by the National Natural Science Foundation of China (Grant Nos. 42277161).

\section*{Reference}


\end{document}